\documentclass{article}

\usepackage[english]{babel}

\usepackage[letterpaper,top=2cm,bottom=2cm,left=3cm,right=3cm,marginparwidth=1.75cm]{geometry}

\usepackage{authblk}
\usepackage{graphicx}
\usepackage{times}
\usepackage[colorlinks=true, allcolors=blue]{hyperref}
\usepackage{amsmath,amssymb,amsfonts}
\usepackage{algorithmic}
\usepackage{subcaption}
\usepackage{textcomp}
\usepackage{hyperref}
\usepackage{array}
\usepackage{rotating}
\usepackage{hhline}
\usepackage{multirow}
\usepackage[table]{xcolor}
\usepackage{colortbl}
\usepackage{booktabs}
\usepackage{float}
\usepackage{placeins}
\usepackage{booktabs}
\usepackage{tabularx}

\title{A Benchmarking Framework for AI models in Automotive Aerodynamics}
\author[1]{Kaustubh Tangsali\thanks{Corresponding Author: Kaustubh Tangsali (\texttt{ktangsali@nvidia.com})}}
\author[1]{Rishikesh Ranade}
\author[1]{Mohammad Amin Nabian}
\author[1]{Alexey Kamenev}
\author[1]{Peter Sharpe}
\author[1]{Neil Ashton}
\author[1]{Ram Cherukuri}
\author[1]{Sanjay Choudhry}

\affil[1]{NVIDIA}

\begin{document}
\definecolor{blue0}{rgb}{1,1,1}
\definecolor{blue1}{rgb}{1,1,1}
\definecolor{blue2}{rgb}{1,1,1}
\definecolor{blue3}{rgb}{1,1,1}
\definecolor{blue4}{rgb}{1,1,1}
\definecolor{blue5}{rgb}{1,1,1}
\definecolor{blue6}{rgb}{1,1,1}
\definecolor{blue7}{rgb}{1,1,1}
\definecolor{blue8}{rgb}{1,1,1}
\definecolor{blue9}{rgb}{1,1,1}
\definecolor{blue10}{rgb}{1,1,1}
\definecolor{blue11}{rgb}{1,1,1}
\definecolor{blue12}{rgb}{1,1,1}
\definecolor{blue13}{rgb}{1,1,1}
\definecolor{blue14}{rgb}{1,1,1}
\definecolor{blue15}{rgb}{1,1,1}
\definecolor{blue16}{rgb}{1,1,1}
\definecolor{blue17}{rgb}{1,1,1}
\definecolor{blue18}{rgb}{1,1,1}
\definecolor{blue19}{rgb}{1,1,1}

\maketitle

\begin{abstract}
In this paper, we introduce a benchmarking framework within the open-source NVIDIA PhysicsNeMo-CFD framework designed to systematically assess the accuracy, performance, scalability, and generalization capabilities of AI models for automotive aerodynamics predictions. The open extensible framework enables incorporation of a diverse set of metrics relevant to the Computer-Aided Engineering (CAE) community. By providing a standardized methodology for comparing AI models, the framework enhances transparency and consistency in performance assessment, with the overarching goal of improving the understanding and development of these models to accelerate research and innovation in the field.
To demonstrate its utility, the framework includes evaluation of both surface and volumetric flow field predictions on three AI models — DoMINO, X-MeshGraphNet, and FIGConvNet — using the DrivAerML dataset . It also includes guidelines for integrating additional models and datasets, making it extensible for physically consistent metrics. This benchmarking study aims to enable researchers and industry professionals in selecting, refining, and advancing AI-driven aerodynamic modeling approaches, ultimately fostering the development of more efficient, accurate, and interpretable solutions in automotive aerodynamics.
\end{abstract}

\section{Introduction}

Computational Fluid Dynamics (CFD) plays an important role in vehicle design to optimize for performance, efficiency and stability. It involves numerically solving the Navier-Stokes equations to calculate flow fields around the vehicle, which may provide key insights into flow dynamics, pressure fluctuations and aerodynamic forces such as drag and lift resulting from changes in the vehicle design. This guidance is crucial for designers and engineers in the design and development phase of these vehicles \cite{hupertz2022}. 

Typical CFD simulations for accurate aerodynamic analysis can take from hours to days depending on the HPC system \cite{hupertz2022} . Turbulence modeling choices further influence the trade-off between computational cost and accuracy. Reynolds Averaged Navier-Stokes (RANS) based methods model time-averaged flow fields using empirically derived turbulence models and embed the influence of all the scales of turbulent motion into the mean flow. As a result, these methods are inexpensive but may be unreliable in resolving complex flow structures such as flow separation, vortex shedding etc \cite{Ashton2018b}. On the other hand, Large Eddy Simulations (LES) or hybrid RANS/LES methods \cite{frohlich2008hybrid} are designed to resolve more and model less of the scales of turbulent motion, thereby providing a more accurate description of flow dynamics. However, these methods can be orders of magnitudes slower than their RANS counterparts. GPU based algorithms are enabling up to an order of magnitude reduction in compute time but are still far from being real-time. 

Recent advancements in AI techniques have led to rapid progress in modeling vehicle aerodynamics. AI-based surrogate models offer the potential for orders-of-magnitude speedups over traditional CFD solvers, significantly accelerating design iterations by providing automotive designers and engineers with faster feedback (typically those best designs will then be validated using traditional CFD solvers). The application of AI in vehicle aerodynamics has garnered significant interest within the research community due to the sheer complexity of the problem, characterized by meshes with millions of elements, intricate geometries, and stringent accuracy requirements.

\begin{sidewaystable}[htb!]
\centering
\renewcommand{\arraystretch}{1.2}
\begin{tabularx}{\textwidth}{
    >{\raggedright\arraybackslash}X
    >{\raggedright\arraybackslash}X
    >{\raggedright\arraybackslash}X
    >{\centering\arraybackslash}X
    >{\raggedright\arraybackslash}X
    >{\raggedright\arraybackslash}X
    >{\raggedright\arraybackslash}X
}
\toprule
\textbf{Model} & \textbf{Scope} & \textbf{Dataset} & \textbf{Samples} & \textbf{Sample Type} & \textbf{Loss Function} & \textbf{Architecture} \\
\midrule
DrivAerNet \cite{elrefaie2024drivaernet} & Predict drag coefficient & DrivAerNet & 4000 & Sedan (DrivAer derivatives) & MSE & RegDGCNN (graph CNN) \\
\hline
Aerodynamics-guided ML \cite{tran2024aerodynamics} & Predict drag coefficient & Proprietary & 500 & SUVs, hatchbacks, sedans, box cars & MSE & PCA-assisted drag-augmented autoencoder \\
\hline
Drag prediction from depth and normal rendering \cite{song2023surrogate} & Predict drag coefficient & CFD on ShapeNet & 9070 & Various vehicle classes & MSE & ResNeXt with attention \\
\hline
DrivAerNet++ \cite{elrefaie2024drivaernet++} & Predict surface fields & DrivAerNet++ & 8000 & Sedan (DrivAer) & MSE & PointNet, RegDGCNN and GCNN \\
\hline
WindsorML-MGN \cite{ashton2024windsorml} & Predict surface fields \& drag coefficient & WindsorML & 355 & Windsor body & MSE & MeshGraphNet \\
\hline
FIGConvNet \cite{choy2025factorized, nvidia2023physicsnemo} & Predict surface fields & DrivAerML & 500 & Sedan (DrivAer derivatives) & MSE \& integral losses & Factorized Implicit Global Convolution Network \\
\hline
3D flow estimation \cite{chen20213d} & Predict volume fields & Proprietary & 1121 & Various vehicle classes (SUVs, sedans, vans, etc.) & MSE, Physics Losses (Continuity) & U-Network \\
\hline
DL for real-time aerodynamic evaluations \cite{jacob2021deep} & Predict drag coefficient \& volume fields & Proprietary & 1000 & Sedan (DrivAer) & Weighted MSE, MAE, etc. & U-Network \\
\hline
DoMINO \cite{ranade2025domino, nvidia2023physicsnemo} & Predict surface \& volume fields (coupled/decoupled) & DrivAerML & 500 & Sedan (DrivAer derivatives) & MSE \& integral losses & Customized DeepONet w/ ball query layers \\
\hline
X-MeshGraphNet \cite{nabian2024x, nvidia2023physicsnemo} & Predict surface \& volume fields (decoupled) & DrivAerML & 500 & Sedan (DrivAer derivatives) & MSE & MeshGraphNet (surface), 3D UNet (volume, both with halo partitioning) \\
\hline
Vehicle flow enrichment \cite{trinh20243d} & Super-resolution of volume flow fields & Proprietary & 1121 & Various vehicle classes (SUVs, sedans, vans, etc.) & Weighted MSE, MAE, Physics Losses (Continuity, Pressure-Poisson) & 3D SR-ResNet (GAN) \\
\hline
Drag guided vehicle image generation \cite{arechiga2023drag} & Generate car shapes while minimizing drag & CFD on ShapeNet & 9070 & Various vehicle classes & MSE & Diffusion model \\
\bottomrule
\end{tabularx}
\caption{A non-exhaustive list of AI models for automotive aerodynamics in the literature.}
\label{table:comparison}
\end{sidewaystable}

\FloatBarrier

Table \ref{table:comparison} presents a non-exhaustive overview of recent AI-driven approaches to automotive aerodynamics. Researchers have applied AI for a range of tasks, including direct prediction of drag coefficients \cite{elrefaie2024drivaernet, song2023surrogate, tran2024aerodynamics}, inferring drag from predicted surface fields \cite{elrefaie2024drivaernet++, ashton2024windsorml, nvidia2023physicsnemo, choy2025factorized}, and reconstructing volume flow fields \cite{ranade2025domino, jacob2021deep, chen20213d, nabian2024x}. All these methods establish mappings from geometric representations—such as signed distance fields—to quantities of aerodynamic interest. Alternative strategies have also emerged, such as the use of generative adversarial networks for flow field super-resolution \cite{trinh20243d}, and the application of AI for shape optimization and design feedback \cite{arechiga2023drag}. The studies leverage a variety of datasets, including DrivAerNet \cite{elrefaie2024drivaernet}, DrivAerNet++ \cite{elrefaie2024drivaernet++}, WindsorML \cite{ashton2024windsorml}, ShapeNet \cite{chang2015shapenetinformationrich3dmodel}, and proprietary data, employing architectures from graph neural networks (e.g., RegDGCNN, MeshGraphNet) to generative models. The development of large-scale, high-quality datasets has been instrumental in advancing model performance. Furthermore, the adoption of diverse loss functions, including physics-informed terms, reflects a trend toward embedding domain knowledge in AI-based aerodynamic modeling. Collectively, these efforts underscore the rapid progress and methodological diversity in this growing research area.

Most AI models designed for automotive aerodynamics are designed for specific tasks, such as prediction of surface fields or aerodynamic forces, and exhibit trade-offs in terms of scalability, accuracy, performance and generalizability. As a result, these models report different error metrics thereby making it difficult to compare them and analyze the different features in these models that make them inherently better at certain tasks. Furthermore, there is inconsistency between the metrics used by engineers and researchers in the CFD community to analyze CFD solver results with those used by the Machine Learning (ML) community in analyzing the AI predictions. Hence, the development of consistent and relevant performance metrics is extremely important to accelerate the development of scalable and accurate AI models with real-world utility and a potential for adoption in the engineering community. 

In this work, we introduce a benchmarking framework for automotive aerodynamics in PhysicsNeMo-CFD, which provides a consistent set of performance metrics to compare and analyze different AI models from a CFD perspective. The framework not only reports $L_2$ errors, surface and volume contour comparisons between ML predictions and CFD solver results but also provides comparison of second level metrics such as aerodynamic forces regression, design trends, line plots, generalization to different meshes etc. In Section \ref{bench_framework}, we describe the various performance metrics included in this utility and provide details on how it can be used to analyze different AI models. In Section \ref{bench_validation}, we introduce the 3 model architectures, DoMINO, X-MeshGraphNet, FIGConvNet that will be evaluated, followed by results and analysis obtained for these models. Finally, we conclude the paper in Section \ref{conclusion} and provide directions for future work. The code repository for PhysicsNeMo-CFD is available at \url{https://github.com/NVIDIA/physicsnemo-cfd}.

\section{PhysicsNeMo-CFD} \label{bench_framework}
NVIDIA PhysicsNeMo-CFD is a sub-module of \href{https://github.com/NVIDIA/physicsnemo/}{NVIDIA PhysicsNeMo framework} that provides the tools needed to integrate pretrained AI models into engineering and CFD workflows. The library is a collection of loosely-coupled workflows around the trained AI models for CFD, with abstractions and relevant data structures.

 PhysicsNeMo-CFD includes benchmarking capabilities that enables evaluation of a wide variety of AI models developed for prediction of automotive aerodynamics. The objective of this work is: 

\begin{itemize}
    \item Equip researchers and engineers with tools to compare AI models using a consistent basis that is important to the CFD community.
    \item Provide a mechanism to understand the strengths and weaknesses of these models, thereby encouraging further research and development.
\end{itemize}
  
As shown in Table \ref{table:comparison}, the various models in this space are primarily designed to predict aerodynamic forces. While some models directly estimate the drag and lift coefficients, others infer surface and volumetric flow fields and subsequently compute the drag coefficient through surface integration of flow quantities. This provides a standardized framework for systematically evaluating these diverse modeling approaches, ensuring consistency and comparability across different methodologies. 

Existing literature predominantly assesses model performance using the coefficient of determination ($R^2$) for drag prediction and point-wise error metrics averaged over mesh elements within the computational domain. These metrics provide a basic sense of accuracy but can mask important deficiencies in aerodynamic predictions. For instance, a model might achieve a high overall $R^2$ yet fail to predict extreme pressure peaks, which are crucial for design decisions. Simularly, when predicting spatially distributed quantities (pressure or velocity fields), pointwise error norms (MSE, RMSE) may not tell the whole story. Designers are often concerned with extreme cases (peak loads, worst-case pressures). To capture this, metrics focusing on the tails of the distribution are useful. Similarly, for drag predictions across many samples, reporting the worst-case error or the number of outliers beyond a certain error threshold provides insight into reliability (this is especially important if using the model in optimization, where an outlier could mislead the design). A comprehensive evaluation should include metrics that capture both global accuracy and specific flow features, as well as quantify uncertainty.  

As a result, traditional metrics alone offer limited insight into model interpretability and are insufficient for evaluating the practical applicability of these models as predictive tools in CAE. Furthermore, variations in dataset partitioning, evaluation criteria, and performance metrics introduce inconsistencies that preclude direct comparisons among different modeling approaches. PhysicsNeMo-CFD addresses these limitations by providing a comprehensive suite of tools designed to enable robust, standardized, and reproducible model evaluation. 

\subsection{Evaluation Metrics}

\subsubsection{Aerodynamic forces} 
Evaluation of aerodynamic forces is critical in automotive aerodynamics, with significant implications on performance, stability, and fuel efficiency. PhysicsNeMo-CFD provides utilities for computing the different components of these forces, including drag, lift, and the corresponding non-dimensional coefficients, facilitating analysis of both surface and volumetric flow predictions around a vehicle. Additionally, it includes post-processing tools such as regression plots, trend analysis, etc. to assess model performance.

While regression plots offer a means to evaluate the goodness-of-fit of AI models, they are not sufficient on their own for aerodynamic assessment. During the vehicle design process, aerodynamicists frequently seek to address "what-if" scenarios, analyzing how design modifications impact aerodynamic performance. In such cases, an AI/ML model's ability to accurately capture relative changes in drag coefficients in response to design variations is crucial. The benchmarking functionality in PhysicsNeMo-CFD provides mechanisms to evaluate models based on these design trends, ensuring robust and practical assessment of predictive accuracy for vehicle design.

\subsubsection{Field comparisons}
In addition to quantitative aerodynamic metrics such as drag and lift coefficients, the visualization of flow fields serves as a critical tool for aerodynamic evaluation. Flow visualizations provide insights into key phenomena, including high- and low-pressure regions, flow separation zones, high-shear areas, and vortical structures, all of which fundamentally influence the aerodynamic forces and provide aerodynamicists with key indicators for design guidance and improvement.

However, analyzing and comparing three-dimensional flow structures presents unique challenges, particularly when dealing with a large set of data samples. In experimental studies, flow field measurements are often available only at specific probe locations, making one-dimensional (1D) line plots a practical choice for comparison. Additionally, two-dimensional (2D) visualizations and surface contours may be available for specific reference configurations. The benchmarking framework addresses these challenges by enabling comparisons of both raw and derived flow fields across diverse output formats, including 1D, 2D, 3D, and 3D manifold representations. Furthermore, it facilitates the simultaneous comparison of multiple samples within a dataset by incorporating statistical averaging and resampling techniques, particularly for cases involving varying geometries.  

\subsubsection{Physics based metrics}
CFD codes typically solve governing equations for mass and momentum conservation to predict the flow around a vehicle. However, direct numerical simulation (DNS) of realistic automotive geometries is computationally prohibitive, necessitating the use of turbulence modeling techniques to approximate turbulent structures. Most CFD simulations strive to satisfy these governing equations as accurately as possible within computational constraints.

In contrast, AI models for fluid flow prediction may or may not explicitly enforce these physical constraints. Many data-driven modeling approaches rely primarily on training data to learn flow behavior, which does not inherently guarantee adherence to the governing equations. To address this limitation, PhysicsNeMo-CFD provides utilities for computing residuals, enabling quantitative assessments of how well AI models satisfy these conservation laws. Such evaluations are crucial for increasing confidence in AI-driven aerodynamic predictions.

However, caution must be exercised when interpreting these residual-based metrics, particularly if the AI/ML model is trained on simulation data that already contain numerical errors. In the absence of explicit physics-based constraints, the best achievable consistency is often limited to the error characteristics of the training data itself.  

\subsubsection{Validation on mesh vs. point cloud}
\label{sec:val-on-mesh-vs-pc}

Meshing is a fundamental prerequisite for engineering simulations, involving the discretization of computational domains into smaller elements where partial differential equations (PDEs) are approximated, and solution fields are computed. In traditional numerical methods, the quality of the mesh significantly influences the accuracy of the computed solutions. However, in applications such as automotive aerodynamics, the meshing process—particularly the generation of a surface mesh and the surrounding computational domain—can account for a substantial portion of the total simulation time. Consequently, meshing often becomes a bottleneck in design iterations, even when solution computation is accelerated using AI models.

In contrast, generating surface and volume point clouds is computationally far less expensive. Therefore, it is critical for AI models to achieve comparable prediction accuracy on sampled point clouds as they do on conventional meshes, ensuring efficiency without compromising predictive performance.

PhysicsNeMo-CFD includes a standardized framework for validating AI models using this metric. It includes a utility for generating a uniformly sampled point cloud directly from STL files, ensuring an even distribution of points across the vehicle surface. The area associated with each point is determined by dividing the total surface area by the number of sampled points, while surface normals are estimated from the corresponding STL facets. Rather than relying on a traditional surface mesh, this uniform point cloud serves as input to AI models for predicting flow field quantities such as pressure and wall-shear stress at each point. Aerodynamic forces are then computed by integrating over the point cloud and compared against ground truth values obtained from high-fidelity simulations on a surface mesh. This approach enables a direct assessment of model performance on point cloud representations.

Currently, this validation metric is designed for surface field evaluations; however, future extensions will incorporate volume field assessments to further enhance its applicability.

\subsection{Including new models and datasets}
\subsubsection{New models}
PhysicsNeMo-CFD provides workflows aimed at offering turn-key access to evaluate different AI models on the DrivAerML dataset for surface and volume predictions. As indicated in \ref{table:comparison}, different models have different prediction scopes; however, most of them predict the surface distributions of quantities like pressure and wall shear stress, and sometimes even the flow around the vehicle. To enable standardized evaluation and comparison, the workflows are designed to take AI model predictions post-processed to .vtp (for surface) and .vtu (for volume) formats as inputs. Generating these .vtp and .vtu files depends on the model's architecture and design, and hence we leave that responsibility to the user; however, we provide some guidance in the GitHub repository on how one can go about generating such files, along with the point clouds (if needed) for the evaluations. The .vtp and .vtu files must contain the true and AI model predicted results on the same points for successful evaluation.

Once the .vtp and .vtu files are generated for the desired test dataset, the workflows can be executed via the command line by passing the appropriate arguments. For a more in-depth "how-to" guide, we recommend readers refer to the \href{https://github.com/NVIDIA/physicsnemo-cfd/tree/main/workflows/bench_example}{GitHub Documentation}. We also provide a Jupyter notebook version of these workflows to enable a better understanding of the underlying evaluation criteria and mechanics.

\begin{figure}[htb!]
    \centering
    \begin{center}
        \includegraphics[width=\textwidth]{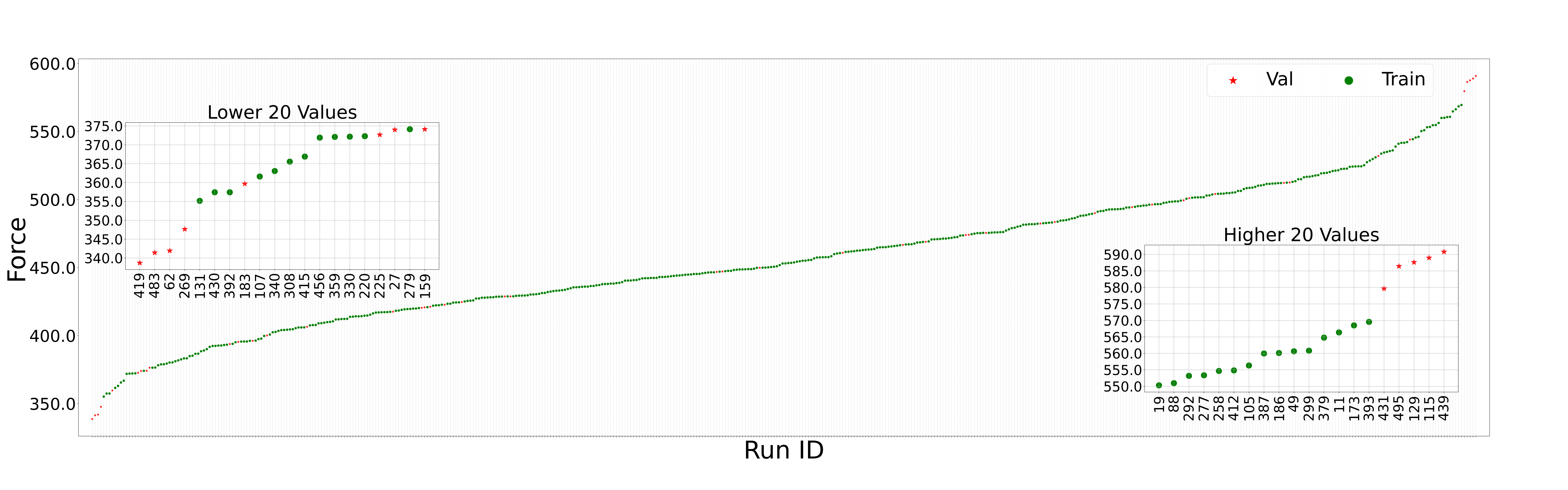}
    \end{center}
    
    \caption{DrivAerML Training-Validation split used for benchmarking}
    \label{drivaer-ml-train-val-split}
\end{figure}

As of this writing, the DrivAerML dataset \cite{ashton2024drivaerml} does not provide guidance on train-validation split. To enable consistent comparison, we propose the split in the benchmarking code. The dataset is split as 90\% for training and 10\% for validation. For selecting the validation set, the entire dataset is sorted by drag force. Then, 10\% of the validation set is chosen to be the top of the sorted dataset, and another 10\% is chosen to be the bottom of the sorted dataset. The rest of the 80\% of the validation set is chosen randomly from the remaining dataset. This allows us to include some out-of-distribution data in the validation set, which is important for the generalizability of the models. Figure~\ref{drivaer-ml-train-val-split} shows the training-validation split pictorially.

Hence, to evaluate any new model using the benchmarking framework on the DrivAerML dataset, one has to ensure the model is trained according to the \href{https://github.com/NVIDIA/physicsnemo-cfd/tree/main/workflows/bench_example/drivaer_ml_files}{train-validation split} proposed, evaluate the trained model on the validation samples, and save the results in .vtp and .vtu files. Finally, use the workflows from PhysicsNeMo-CFD to make the comparisons.

\subsubsection{New datasets}
The workflows from PhysicsNeMo-CFD are customizable to enable comparisons for different CFD datasets. The steps required to test the model on new datasets are similar to the steps mentioned in the previous section. The different parameters of the new dataset corresponding to variable names, data paths etc. can be configured through a configuration file.

\begin{figure}[htbp]
  \centering
  \begin{subfigure}{\textwidth}
    \centering
    \includegraphics[width=0.9\linewidth]{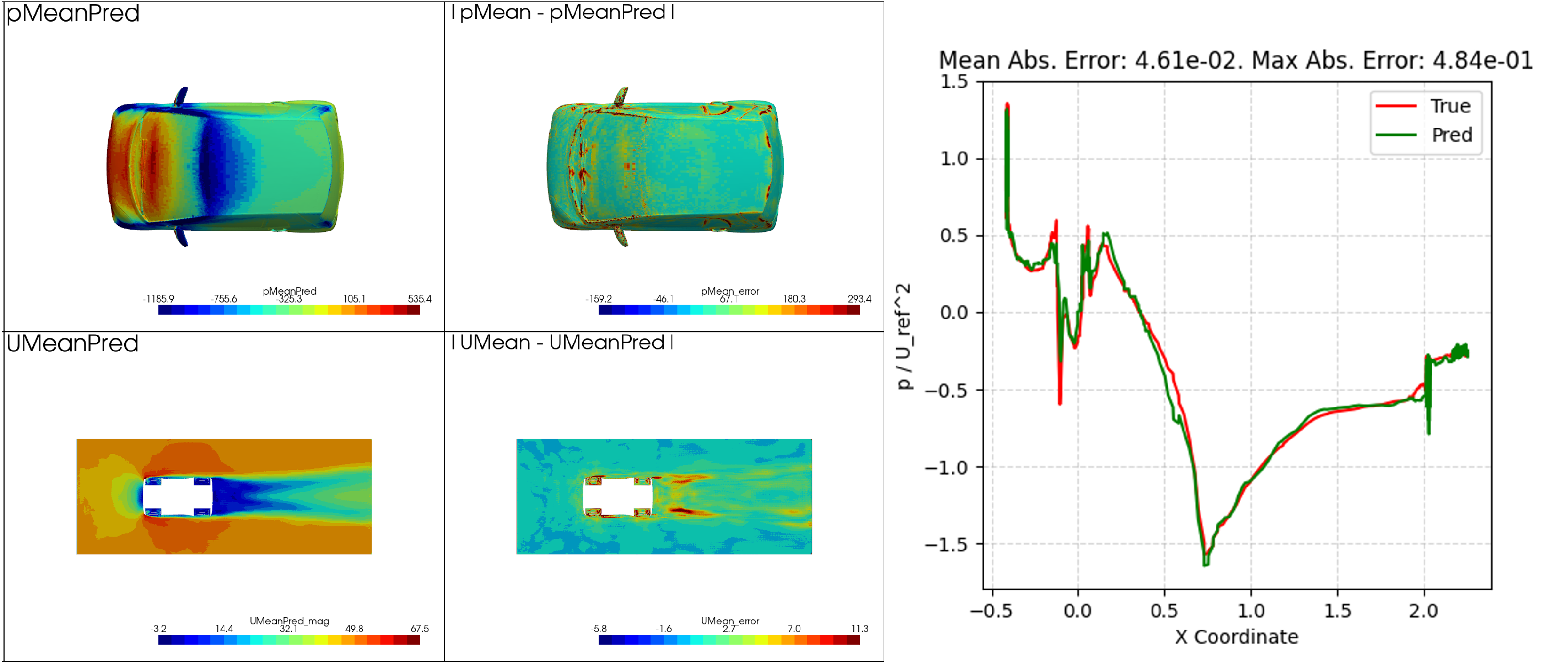}
    \caption{DriveSim dataset, model used DoMINO}
    \label{fig:image1}
  \end{subfigure}

  \vspace{0.5cm} 

  \begin{subfigure}{\textwidth}
    \centering
    \includegraphics[width=0.9\linewidth]{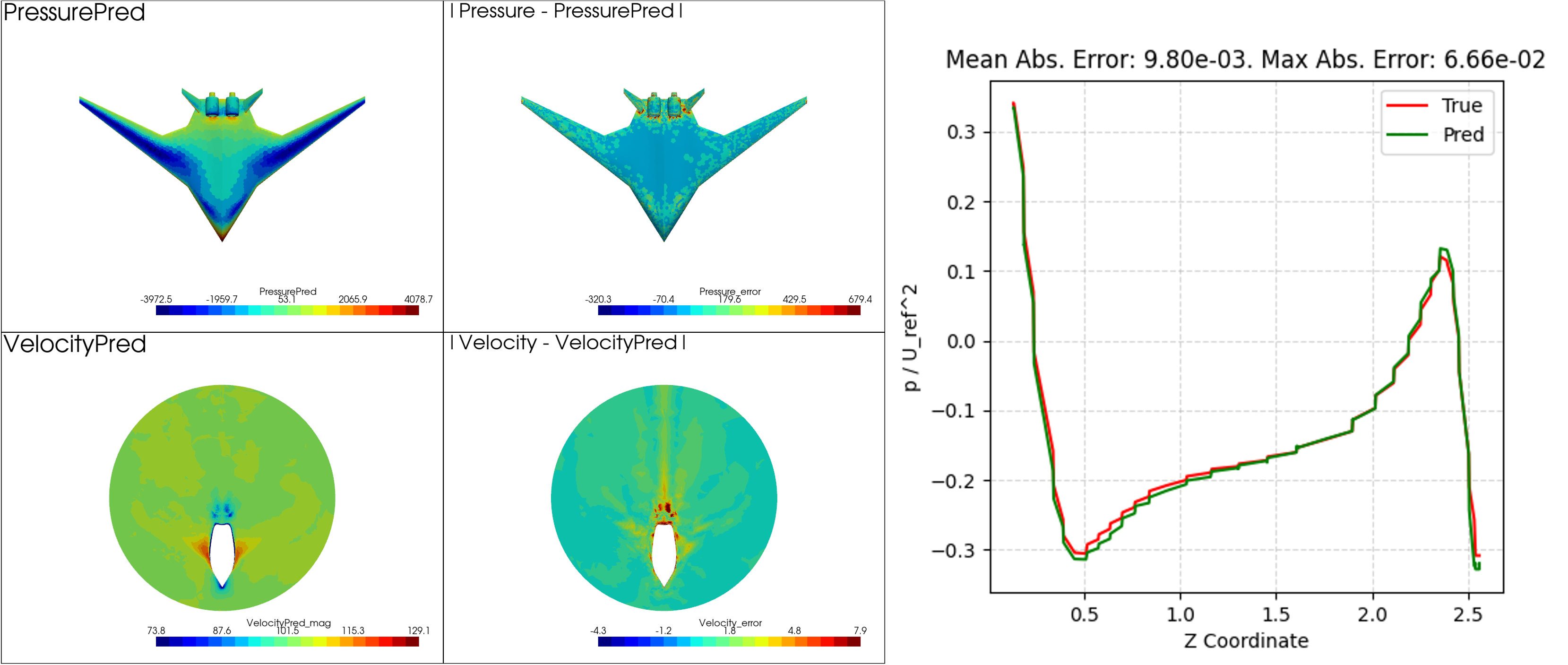}
    \caption{Aircraft dataset, model used DoMINO}
    \label{fig:image2}
  \end{subfigure}

  \caption{Using PhysicsNeMo-CFD for evaluating AI models trained on new datasets. Figures show surface contours, volume contours and 1D line plots of normalized surface pressure (along vehicle centerline) of DoMINO model output. DoMINO model was trained on two datasets, one on a passenger car dataset (DriveSim) and one on an Aircraft dataset. }
  \label{fig:new-datasets}
\end{figure}

We have tested the benchmarking setup across several other datasets such as DriveSim (an internally generated RANS dataset for passenger cars using OpenFOAM), DriveSim+ (an internally generated RANS dataset for passenger cars with a commercial solver), and many other datasets, some including external aerodynamics of aircraft (Figure~\ref{fig:new-datasets}). Additionally, while the workflows from PhysicsNeMo-CFD are designed to be used for ML model benchmarking, they can also be used more generally for any CFD model verification and comparison.

\section{Validation} \label{bench_validation}

Next, the benchmarking framework is validated on select model architectures and details. This section provides details of the model architectures, datasets and analyzes the results obtained across different metrics available in PhysicsNeMo-CFD.

\subsection{Model architectures} \label{model_archs}

\subsubsection{DoMINO}

\begin{figure}[htb!]
    \centering
    \begin{center}
        \includegraphics[scale=0.5]{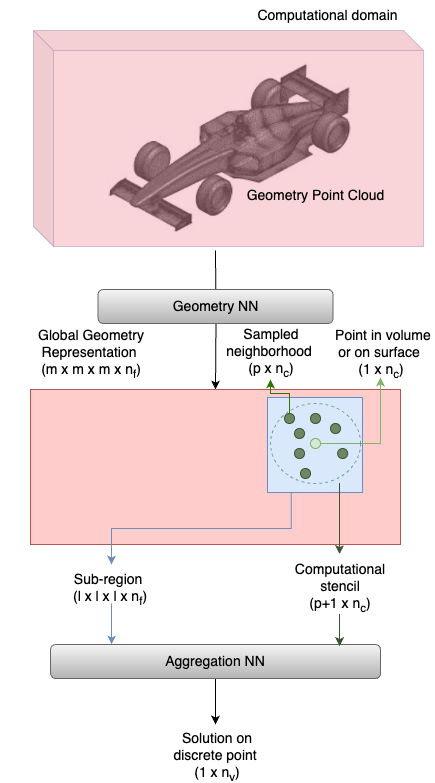}
    \end{center}
    
    \caption{DoMINO model overview.}
    \label{fig_domino}
\end{figure}

The DoMINO model is a neural operator based model which uses elements of the DeepONet architecture \cite{lu2019deeponet}. It learns local geometry encodings from point cloud representations to predict PDE solutions on discrete points sampled in computational domain using dynamically constructed computational stencils in local regions around it. The model leverages local features within sub-regions of the computational domain to predict solutions on both the surface of car as well as in the computational domain volume around it. Predicting both sets of quantities is extremely important for many applications in engineering simulations. An overview diagram for the DoMINO model architectures is shown in \ref{fig_domino}. However, a detailed description of the model architecture can be found in the paper \cite{ranade2025domino}.

A 3D bounding box is constructed around the geometry to represent a computational domain (light red) that captures the most important solution fields required for design guidance. The computational domain is discretized with a fixed structured representation of resolution, $m_x$ * $m_y$ * $m_z$, where $m_x$, $m_y$, $m_z$ are resolutions in x, y, z directions respectively. The geometry point cloud is projected onto this grid using a Geometry NN, which consists of trainable, dynamic point convolution kernels. 

Next, we sample a batch of discrete points (shown in light green) in the computational domain where we want to evaluate the solution. For each of the sampled points in a batch, a subregion (shown in blue) of size $l_x$ * $l_y$ * $l_z$ is defined around it in the computational domain. Trainable point convolutional kernels are used to extract features from global geometry encoding in this local region defined around each point in a batch. The extracted features are further processed into a lower-dimensional vector of length, $\eta_l$, using fully connected neural networks. 

After the local geometry encoding is computed, it is used to predict the solution on the light-green sampled point. For this, we dynamically sample a set of $p$ neighboring points around each sampled point. These input features on each of the p+1 points concatenated with the local geometry encoding, $\eta_l$, and processed using a fully connected network known as the aggregation network. The output of the aggregation network is the solution vector on each of these p+1 points. The final solution on the light-green points is calculated by inverse-distance averaging of the p+1 points in the computational stencil. 

\subsubsection{FIGConvNet}
Factorized Implicit Global Convolution Network (FIGConvNet) is a novel architecture that efficiently solves CFD problems for large 3D meshes with arbitrary input and output geometries. FIGConvNet achieves quadratic complexity $O(N^2)$, a significant improvement over existing 3D neural CFD models that require cubic complexity $O(N^3)$. The model approach combines Factorized Implicit Grids to approximate high-resolution domains, efficient global convolutions through 2D reparameterization, and a U-shaped architecture for effective information gathering and integration. A detailed description of the model architecture can be found in the paper \cite{choy2025factorized}.

\subsubsection{X-MeshGraphNet}
X-MeshGraphNet is a collection of AI models designed to predict aerodynamic quantities at scale with high accuracy and efficiency, mainly targeting the scalability challenges of AI models. It comprises two complementary surface and volume models. Both of these models rely on a novel partitioning scheme with halo regions and gradient aggregation to improve scalability and enabling working with extremely large meshes and voxel grids.

The surface model is designed to predict aerodynamic properties such as pressure and wall shear stresses on vehicle surfaces using a scalable multi-scale graph neural network architecture. It generates custom graphs directly from CAD models, eliminating the need for pre-generated simulation meshes. The process begins by uniformly sampling points from the vehicle's surface to create a point cloud, which is then connected using k-nearest neighbor (k-NN) graph construction. To capture both global and local aerodynamic phenomena, multi-scale graphs are built, where coarser graphs represent large-scale structures and finer graphs refine local details. The model employs the X-MeshGraphNet architecture \cite{nabian2024x} with a message-passing mechanism that propagates information across graph nodes, each representing a surface point with features such as position, surface normal, and Fourier-transformed coordinates. A partitioning scheme with halo regions enables seamless message passing between subgraphs, supporting large-scale geometries while ensuring memory efficiency through techniques like activation checkpointing and mixed-precision training. This predicts aerodynamic targets such as surface pressure and wall shear stresses using a mean squared error (MSE) loss. Validated on the DrivAerML dataset, it has demonstrated high accuracy, closely matching CFD results while significantly reducing computational costs.

\begin{figure*}[htbp]
    \centering
    \begin{subfigure}{0.49\textwidth}  
        \centering
        \includegraphics[width=\textwidth]{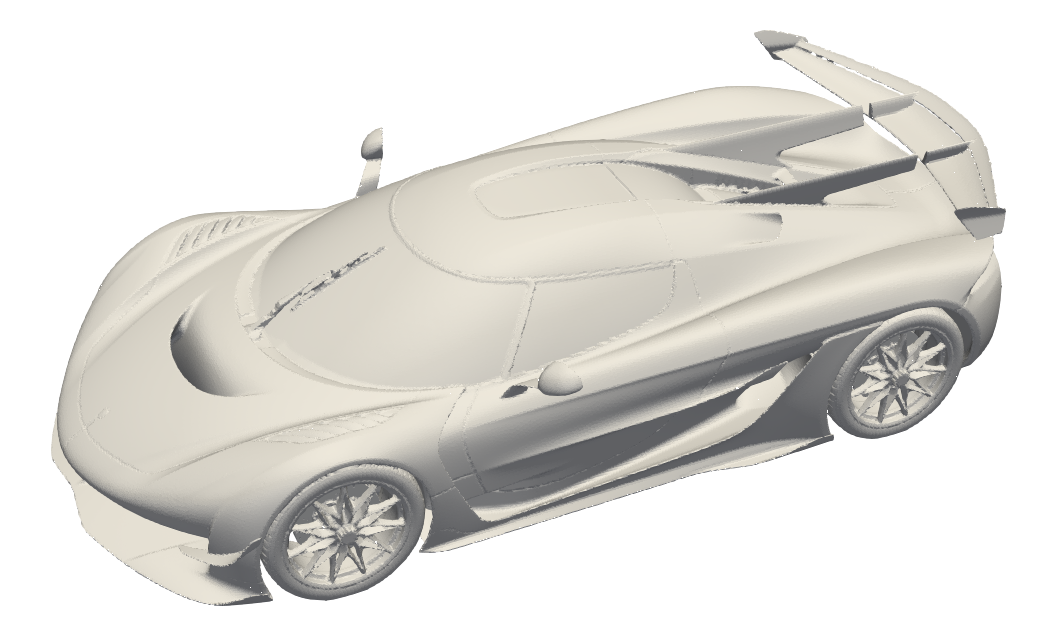}
        \caption{Original representation.}
        \label{fig:sub1}
    \end{subfigure}
    \hfill
    \begin{subfigure}{0.49\textwidth}  
        \centering
        \includegraphics[width=\textwidth]{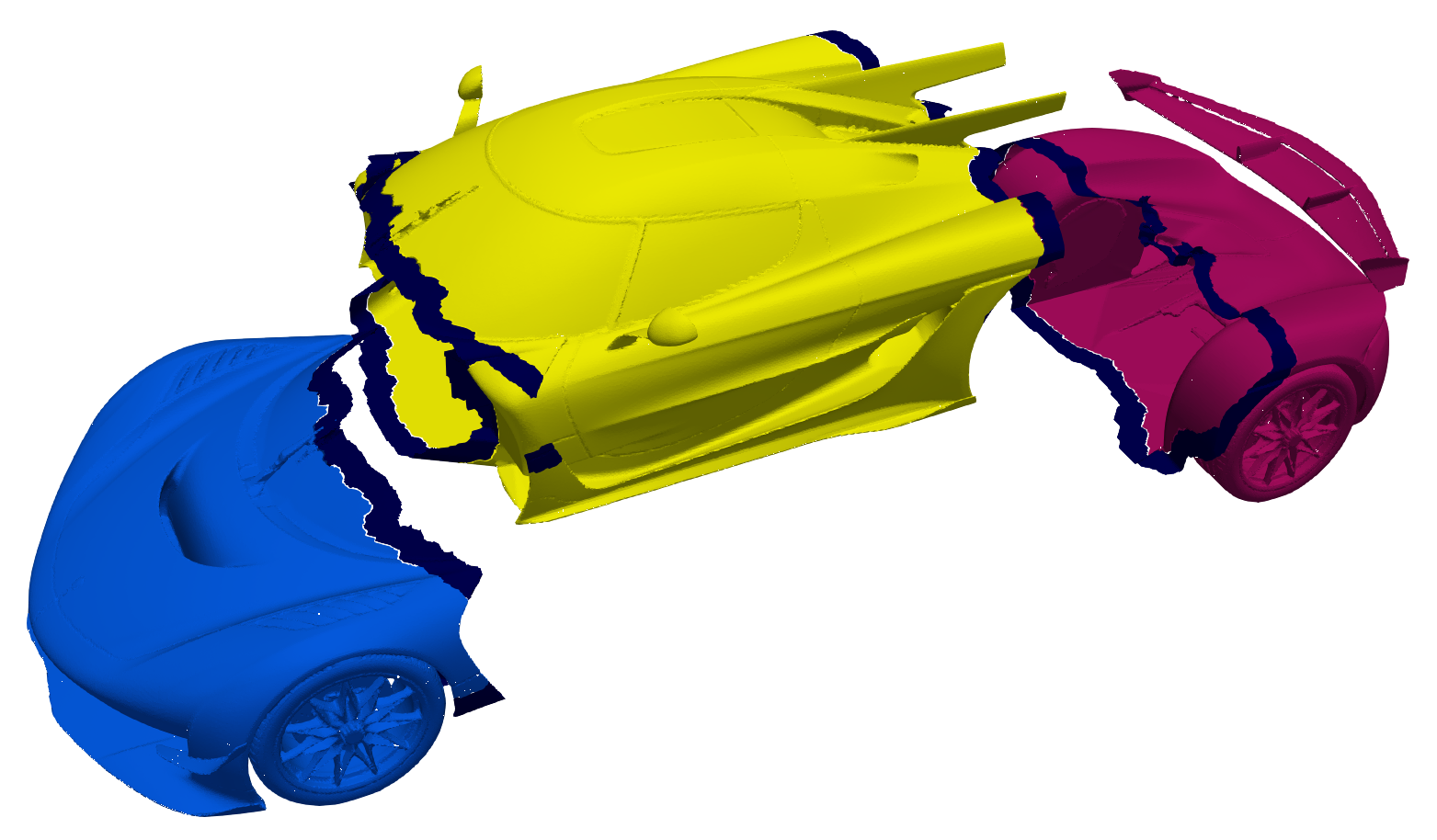}
        \caption{Partitioned representation with Halo regions.}
        \label{fig:sub2}
    \end{subfigure}
    \caption{Illustration of the X-MeshGraphNet partitioning scheme with Halo on a Koenigsegg car.}
    \label{fig:partitioning}
\end{figure*}

The volume model extends the halo partitioning with gradient aggregation scheme to CNNs and UNet-like models, and is designed to predict aerodynamic properties such as pressure and velocity fields around vehicles using a scalable 3-level UNet architecture. It operates on a voxelized representation of the volumetric domain, where a bounding box is defined around the vehicle and discretized into a uniform voxel grid. The vehicle's CAD geometry is embedded by marking surface-intersecting voxels, enabling precise aerodynamic property prediction. The model leverages a hierarchical UNet structure, progressively downsampling and upsampling the voxel grid to capture both global flow patterns and local aerodynamic details. Skip connections preserve fine-grained spatial information, while a partitioning scheme with halo regions ensures seamless data flow across subdomains. During training, gradients from all partitions are aggregated for consistency, supporting large-scale aerodynamic simulations. Input features include voxel center coordinates, surface proximity encoding, and Fourier features for spatial periodicity. This model predicts pressure and 3D velocity components $(u,v,w)$, employing advanced training optimizations such as mixed-precision training and activation checkpointing to manage memory and computational costs efficiently. Its ability to handle large volumetric domains while accurately capturing flow dynamics makes it a powerful tool for automotive aerodynamic evaluation, complementing its surface-based counterpart.

\subsection{Datasets}

\subsubsection{DrivAerML Dataset}
The DrivAerML dataset \cite{ashton2024drivaerml} is a high-fidelity, open-source dataset designed to advance ML applications in automotive aerodynamics. It includes 500 parametrically varied geometries based on the widely used DrivAer notchback vehicle model, generated using hybrid RANS-LES (HRLES) simulations. This approach captures complex aerodynamic phenomena such as turbulent wakes, flow separation, and aerodynamic drag with industry-standard accuracy. The dataset creation process involved automated meshing, scale-resolving CFD simulations, and comprehensive post-processing to ensure consistent, high-quality data. Each simulation includes volumetric flow fields, surface data, and aerodynamic force coefficients. The dataset is publicly available under a CC-BY-SA license.

DrivAerML bridges a critical gap in the ML and automotive aerodynamics communities by providing realistic, high-resolution CFD data at a scale previously unavailable. It covers diverse vehicle design variations using 16 geometric parameters, generating meaningful aerodynamic responses that span a wide range of drag and lift coefficients. The simulations were validated against experimental wind tunnel data, demonstrating strong agreement in flow structures, pressure distributions, and wake dynamics. With its large-scale, physics-rich structure, DrivAerML supports research into ML-driven surrogate modeling, optimization, and real-time aerodynamic predictions. This initiative sets a new benchmark for open-source aerodynamic datasets, offering a robust foundation for developing advanced ML algorithms tailored to automotive design challenges.

\subsubsection{Other Datasets}
The underlying metrics and workflows from PhysicsNeMo-CFD can be leveraged to compare models on any datasets for external aerodynamics, however, we have only tested with DrivAerML dataset and includes a full integration, while for other datasets, we had done limited testing and are not fully integrated. An overview of additional publicly available automotive aerodynamic datasets is provided below.

\paragraph{WindsorML} \cite{ashton2024windsorml}:
WindsorML is a large-scale, high-fidelity dataset featuring 355 geometric variants of the Windsor body. Each variant was simulated using Wall-Modeled Large-Eddy Simulations (WMLES) with over 280 million computational cells, employing a Cartesian immersed-boundary method for precise aerodynamic predictions. The dataset includes time-averaged 3D volumetric flow fields, surface boundary data, aerodynamic force and moment coefficients, and parameterized CAD models defined by seven geometric variables relevant to automotive design. All simulations were conducted under fixed flow conditions. The dataset is publicly available under a CC-BY-SA license.

\paragraph{AhmedML} \cite{ashton2024ahmedml}:
AhmedML is a large-scale, open-source dataset featuring 500 geometric variants of the Ahmed car body. Each variant was simulated using a high-fidelity, time-accurate hybrid RANS-LES approach with OpenFOAM, employing 15-20 million prismatic and hex-dominant cells per simulation. The dataset includes time-averaged volumetric flow fields, surface boundary data, aerodynamic force and moment coefficients, geometric CAD models, and 2D slices of key flow variables. STL files and a complete OpenFOAM case setup are provided for reproducibility. Flow conditions match experimental benchmarks, and the dataset captures essential aerodynamic features such as 3D vortical structures, pressure-induced separation, and rear-slant flow reattachment. Validation against experimental data demonstrated strong correlation for drag and lift coefficients. The dataset is publicly available under a CC-BY-SA license.

\paragraph{DrivAerNet++} \cite{elrefaie2024drivaernet}:
DrivAerNet++ is a comprehensive, multimodal dataset featuring 8,000 high-fidelity CFD-simulated car designs with diverse configurations, including fastback, notchback, and estateback models, along with various underbody and wheel designs for both ICE and EV platforms. Each entry includes detailed 3D meshes, aerodynamic coefficients, parametric models, and annotated car components, supporting tasks such as design optimization, CFD simulation acceleration, and aerodynamic performance prediction. The dataset provides multiple data representations, including 3D point clouds, parametric data, and segmented surface fields, totaling over 39 TB of publicly accessible engineering data. DrivAerNet++ is released under the CC BY-NC 4.0 license.

\subsection{Experimental setup} \label{expt_setup}

In this section we provide details related to the training and evaluation of the 3 AI models, DoMINO, X-MeshGraphNet and FIGConvNet. The models are trained and evaluated with the same data split. Of the 484 samples available in the DrivAerML dataset, 436 are reserved for training and 48 for testing. The training and testing ids of consistent for the 3 models. The X-MeshGraphNet and FIGConvNet are trained to evaluate only the surface fields, while the DoMINO is trained to evaluate both the surface and volume fields using a coupled model (a single model predicting both types of fields).

In the case of X-MeshGraphNet and FIGConvNet a uniform point cloud of 4.5 million points are sampled on the surface of the vehicle from the STL and solutions from the simulation surface mesh are linearly interpolated on this point cloud. Additionally the connectivity information is also computed in X-MeshGraphNet as it requires a graph-based input data structure. For evaluation, the same procedure is followed given an STL, a uniform point cloud (4.5 M points for X-MeshGraphNet and 500 K points for FIGConvNet) is sampled on the surface. Surface fields evaluated by the models on the point clouds are linearly interpolated back on to the simulation mesh for visualization and benchmarking. The DoMINO model is trained on the simulation mesh points and evaluated on the same. The model architectures, configurations, training and inference pipelines are published in the \href{https://github.com/NVIDIA/physicsnemo/tree/main/examples/cfd/external_aerodynamics}{PhysicsNeMo GitHub repo}.

During the inference stage, the model predictions are written back to the raw simulation files under different variable names. These raw files with both the ground truth simulation results as well as the prediction results are used to calculate the different metrics described in Section \ref{bench_framework}.

\subsection{Results and discussions}

In this section, we present the different results evaluated using the the benchmarking framework for a range of metrics. As described in Section \ref{expt_setup}, the results are evaluated over 48 validation samples. It must be noted that the validation set consists of both in-distribution and out-of-distribution samples. The out-of-distribution samples contain cases with some of the lowest and highest drag forces and are not seen by the models during training. For the purposes of contour comparisons, we use sample IDs 419 and 439 as representative examples because they correspond to samples with the smallest and highest drag in the entire dataset. 

\subsubsection{Surface results}

Results from surface benchmarking on the validation set for X-MeshGraphNet, FIGConvNet and DoMINO are reported below.

\paragraph{Average error metrics} Table~\ref{table:l2-errors} and Table~\ref{table:area-wt-l2-errors} show the L2 errors and area-weighted L2 errors evaluated on the cell centers of the mesh, respectively. Based on L2 errors, the results of X-MeshGraphNet and DoMINO are close, while based on area-weighted L2 errors (L2 error at each cell center is multiplied by the area of that cell), DoMINO performs significantly better than the other models. This is likely because DoMINO trains on the surface mesh data (data from cell centers) directly, while X-MeshGraphNet and FIGConvNet both use solutions interpolated on an uniform point cloud during training.

\begin{table}[H]
\centering
\caption{L2 errors on validation samples (Lower is better)}
\label{table:l2-errors}
\begin{tabular}{lrrr}
\toprule
 & X-MeshGraphNet & FIGConvNet & DoMINO \\
\midrule
Pressure & \cellcolor{blue11}0.14 & \cellcolor{blue0}0.21 & \cellcolor{blue19}0.10 \\
Wall Shear Stress (x) & \cellcolor{blue19}0.17 & \cellcolor{blue0}0.32 & \cellcolor{blue17}0.18 \\
Wall Shear Stress (y) & \cellcolor{blue19}0.22 & \cellcolor{blue0}0.62 & \cellcolor{blue17}0.26 \\
Wall Shear Stress (z) & \cellcolor{blue18}0.29 & \cellcolor{blue0}0.53 & \cellcolor{blue19}0.28 \\
\bottomrule
\end{tabular}
\end{table}

\begin{table}[H]
\centering
\caption{Area weighted L2 errors on validation samples (Lower is better)}
\label{table:area-wt-l2-errors}
\begin{tabular}{lrrr}
\toprule
 & X-MeshGraphNet & FIGConvNet & DoMINO \\
\midrule
Pressure & \cellcolor{blue0}0.14 & \cellcolor{blue1}0.14 & \cellcolor{blue19}0.08 \\
Wall Shear Stress (x) & \cellcolor{blue8}0.13 & \cellcolor{blue0}0.16 & \cellcolor{blue19}0.10 \\
Wall Shear Stress (y) & \cellcolor{blue16}0.24 & \cellcolor{blue0}0.34 & \cellcolor{blue19}0.23 \\
Wall Shear Stress (z) & \cellcolor{blue8}0.25 & \cellcolor{blue0}0.29 & \cellcolor{blue19}0.19 \\
\bottomrule
\end{tabular}
\end{table}

\begin{figure}[ht!]
     \centering
     \begin{subfigure}[b]{0.48\textwidth}
         \centering
         \subcaption{DoMINO trends}
         \includegraphics[width=\textwidth]{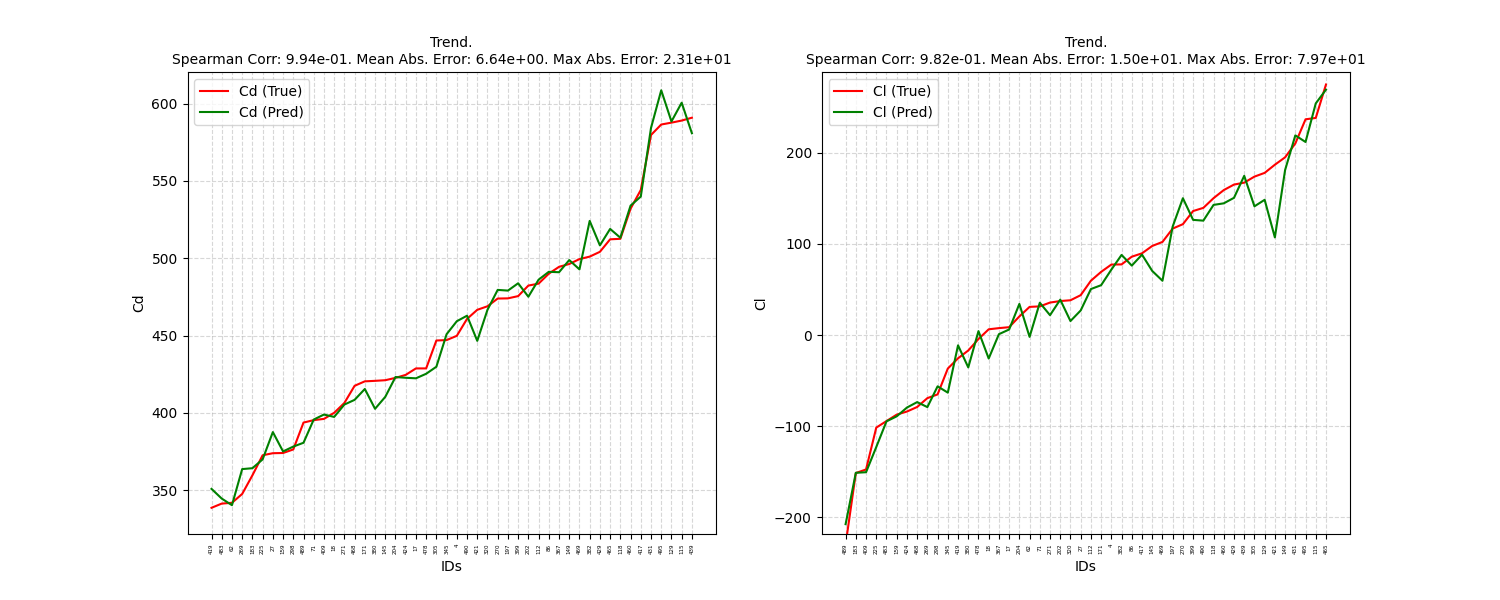}
         \label{fig:y equals x 4}
     \end{subfigure}
     \hfill
     \begin{subfigure}[b]{0.48\textwidth}
         \centering
         \subcaption{DoMINO regression}
         \includegraphics[trim={1cm 1cm 1cm 1cm},clip,width=\textwidth]{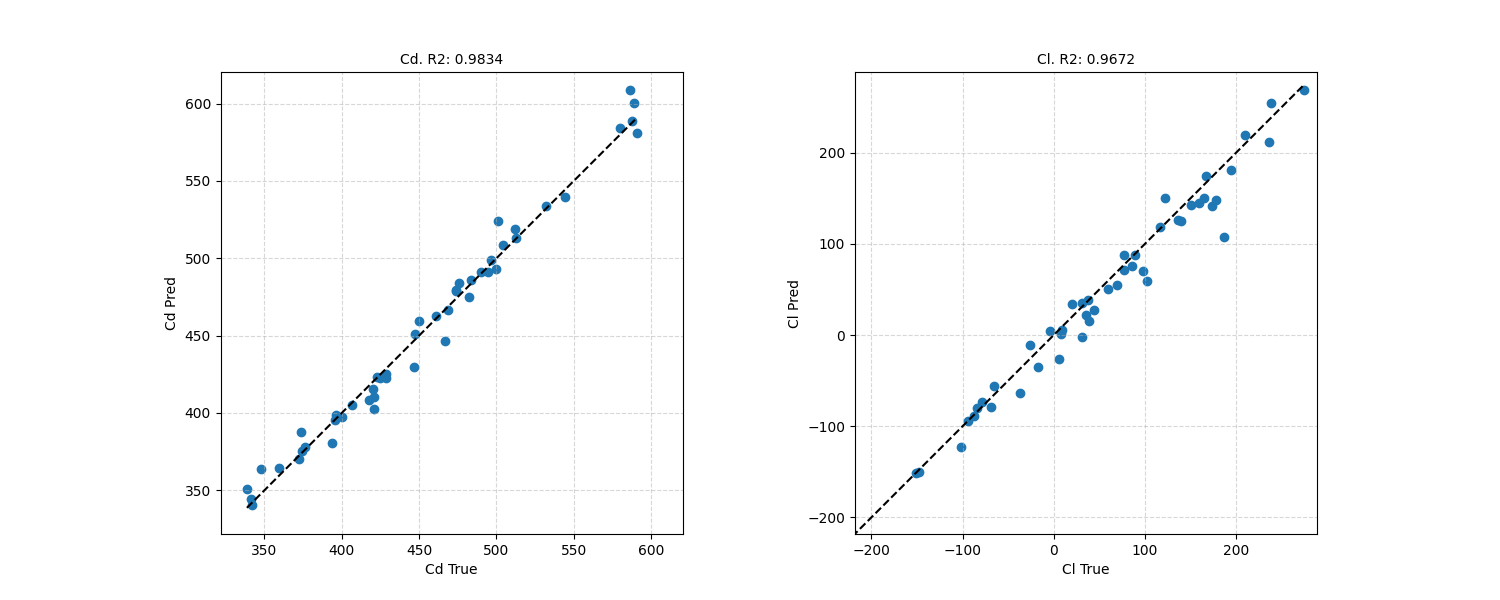}
         \label{fig:y equals x 5}
     \end{subfigure}
     \vfill
     \begin{subfigure}[b]{0.48\textwidth}
         \centering
         \subcaption{FIGConvNet trends}
         \includegraphics[width=\textwidth]{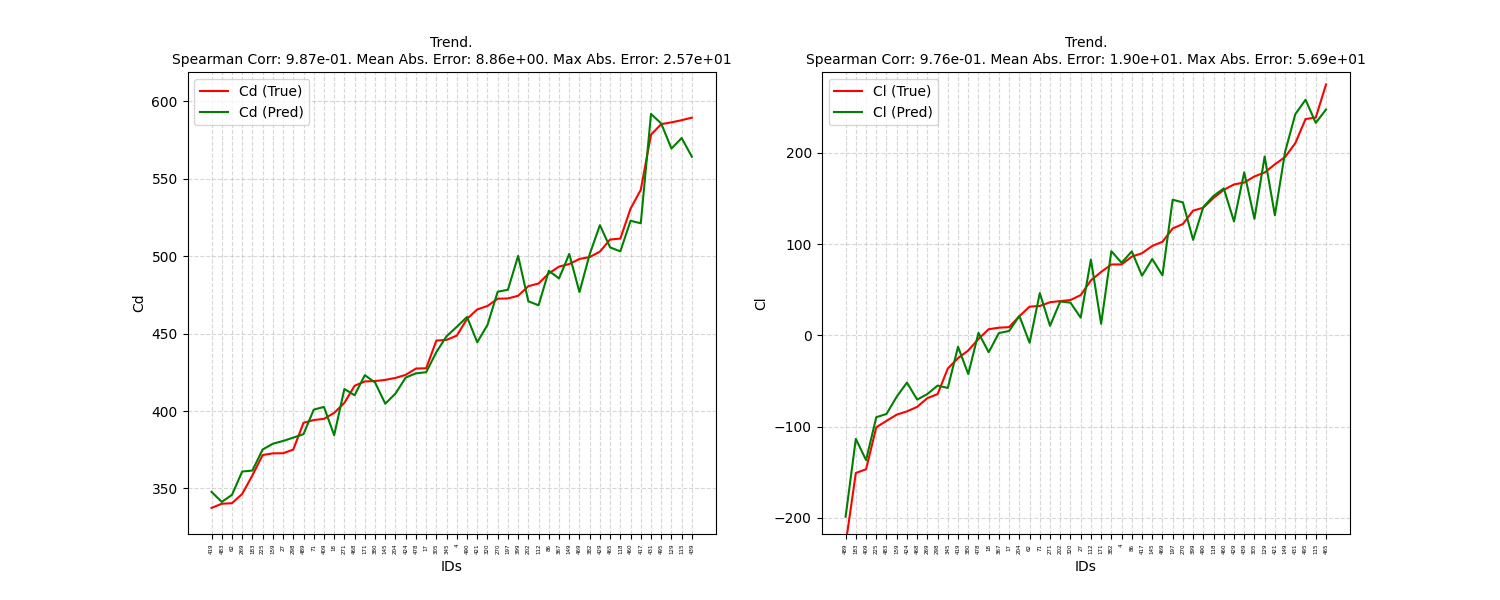}
         \label{fig:three sin x 5}
     \end{subfigure}
     \hfill
     \begin{subfigure}[b]{0.48\textwidth}
         \centering
         \subcaption{FIGConvNet regression}
         \includegraphics[trim={1cm 1cm 1cm 1cm},clip,width=\textwidth]{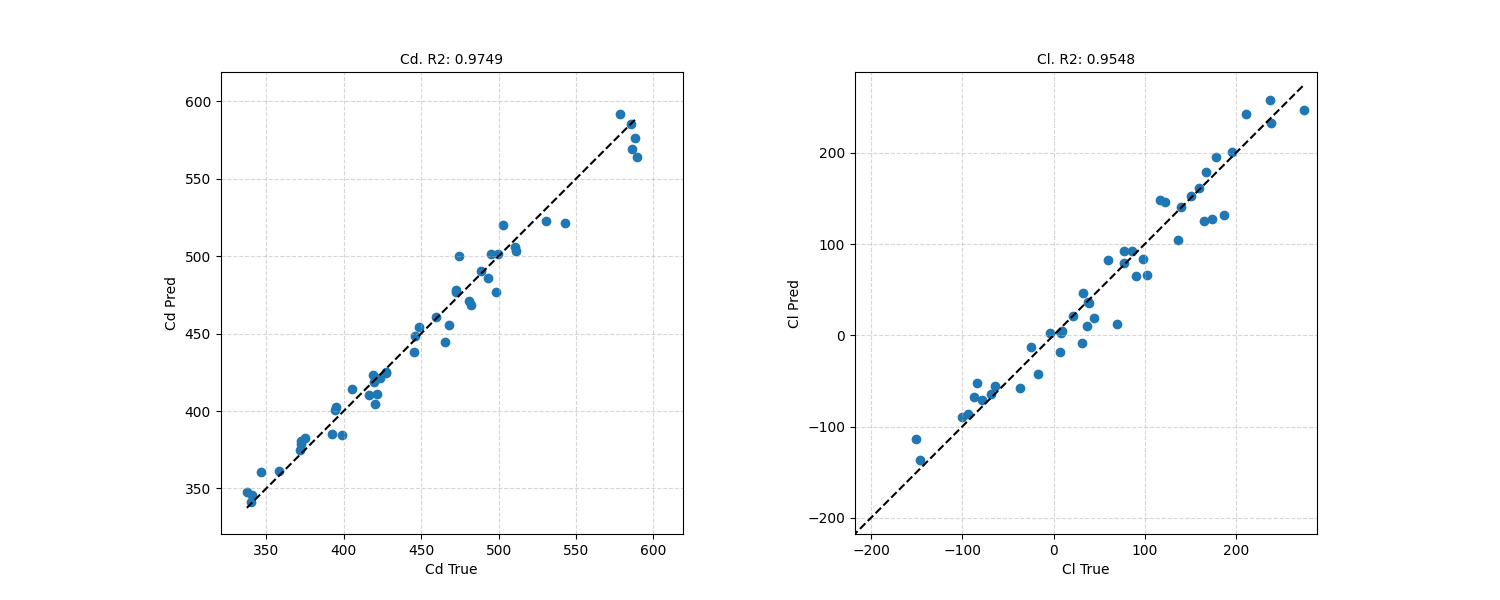}
         \label{fig:three sin x 6}
     \end{subfigure}
     \vfill
     \begin{subfigure}[b]{0.48\textwidth}
         \centering
         \subcaption{X-MeshGraphNet trends}
         \includegraphics[width=\textwidth]{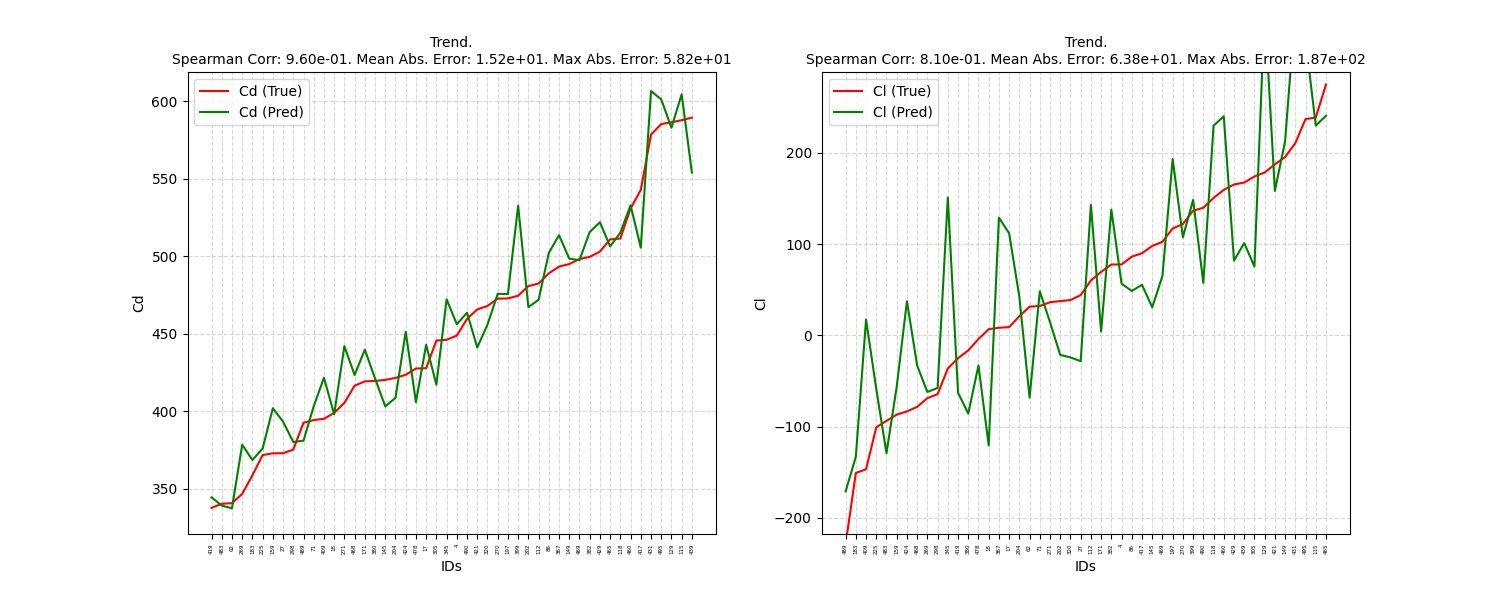}
         \label{fig:five over x 1}
     \end{subfigure}
     \hfill
     \begin{subfigure}[b]{0.48\textwidth}
         \centering
         \subcaption{X-MeshGraphNet regression}
         \includegraphics[trim={1cm 1cm 1cm 1cm},clip,width=\textwidth]{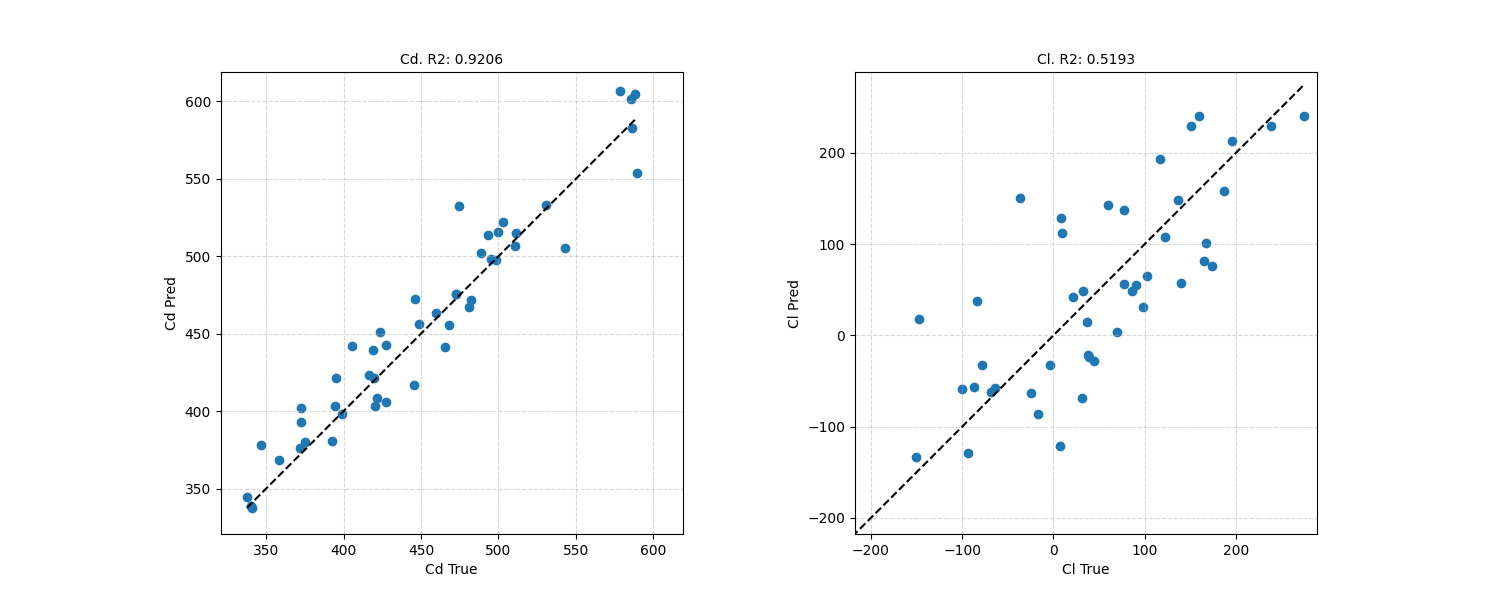}
         \label{fig:five over x 2}
     \end{subfigure}     
     
    \caption{Drag and lift force design trends and regression comparing predicted and simulated ground truth}
    \label{design_trends}
\end{figure}

\paragraph{Aerodynamic forces trends and regression} Figure~\ref{design_trends} shows qualitative comparisons between the design trends of the three models, while Table~\ref{table:trend-spearman} and Table~\ref{table:trend-errors} provide a quantitative comparison. The aerodynamic forces are calculated by integrating the predicted field quantities over the car surface and compared with those calculated from ground truth fields. We use a combination of the Spearman coefficient and absolute differences to analyze how well the model captures the original design trend, which is plotted by ranking the validation samples in ascending order with respect to the chosen metric (drag/lift force). Overall, we find that FIGConvNet and DoMINO perform better based on these metrics compared to X-MeshGraphNet, with DoMINO performing slightly better than FIGConvNet. This is likely because both FIGConvNet and DoMINO accurately predict solutions on surfaces of the car that have a huge impact on aerodynamic forces. These correspond to regions on the car surface, such as bumper, wheels etc., where the magnitude of the area-weighted pressure in the flow direction is relatively high. Additionally, X-MeshGraphNet solely focuses on predicting the raw variables, in contrast to FIGConvNet and DoMINO, which also have a loss term for the integrated quantities.

Additionally, Figure~\ref{design_trends} and Table~\ref{table:r2-scores} show the qualitative and quantitative comparisons of the regression on drag and lift forces. Similar to the previous trend analysis, FIGConvNet and DoMINO perform better than X-MeshGraphNet, with DoMINO performing the best among the three. The reasons for the performance are likely the same as described above.

\begin{table}[H]
\centering
\caption{Trend analysis (Spearman coefficient) on validation samples (Higher is better)}
\label{table:trend-spearman}
\begin{tabular}{lrrr}
\toprule
 & X-MeshGraphNet & FIGConvNet & DoMINO \\
\midrule
Spearman Coeff. Drag & \cellcolor{blue0}0.96 & \cellcolor{blue15}0.99 & \cellcolor{blue19}0.99 \\
Spearman Coeff. Lift & \cellcolor{blue0}0.81 & \cellcolor{blue18}0.98 & \cellcolor{blue19}0.98 \\
\bottomrule
\end{tabular}
\end{table}

\begin{table}[H]
\centering
\caption{Trend analysis (Errors) on validation samples (Lower is better)}
\label{table:trend-errors}
\begin{tabular}{lrrr}
\toprule
 & X-MeshGraphNet & FIGConvNet & DoMINO \\
\midrule
Mean Abs. Error Drag & \cellcolor{blue0}15.23 & \cellcolor{blue14}8.86 & \cellcolor{blue19}6.64 \\
Max Abs. Error Drag & \cellcolor{blue0}58.16 & \cellcolor{blue17}25.72 & \cellcolor{blue19}23.08 \\
Mean Abs. Error Lift & \cellcolor{blue0}63.75 & \cellcolor{blue17}19.00 & \cellcolor{blue19}15.04 \\
Max Abs. Error Lift & \cellcolor{blue0}187.42 & \cellcolor{blue19}56.90 & \cellcolor{blue15}79.71 \\
\bottomrule
\end{tabular}
\end{table}

\begin{table}[H]
\centering
\caption{R2 scores on validation samples (Higher is better)}
\label{table:r2-scores}
\begin{tabular}{lrrr}
\toprule
 & X-MeshGraphNet & FIGConvNet & DoMINO \\
\midrule
R2 Drag & \cellcolor{blue0}0.92 & \cellcolor{blue16}0.97 & \cellcolor{blue19}0.98 \\
R2 Lift & \cellcolor{blue0}0.52 & \cellcolor{blue18}0.95 & \cellcolor{blue19}0.97 \\
\bottomrule
\end{tabular}
\end{table}

\begin{figure}[ht!]
     \centering
     \begin{subfigure}[b]{0.48\textwidth}
         \centering
         \subcaption{DoMINO Top Centerline}
         \includegraphics[trim={1cm 1cm 1cm 0.75cm},clip,width=\textwidth]{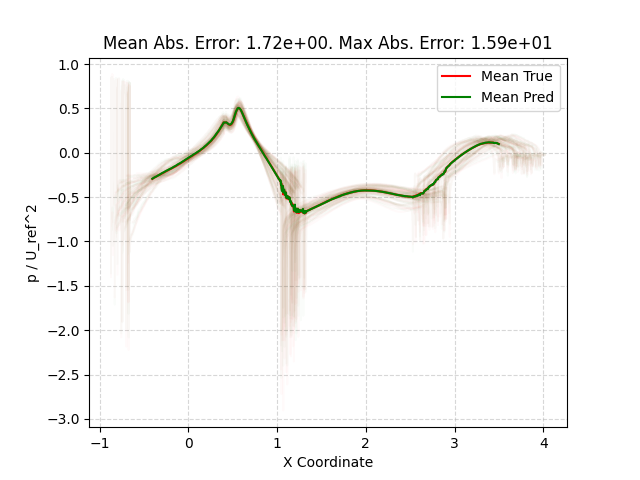}
         \label{fig:y equals x 6}
     \end{subfigure}
     \hfill
     \begin{subfigure}[b]{0.48\textwidth}
         \centering
         \subcaption{DoMINO Bottom Centerline}
         \includegraphics[trim={1cm 1cm 1cm 1cm},clip,width=\textwidth]{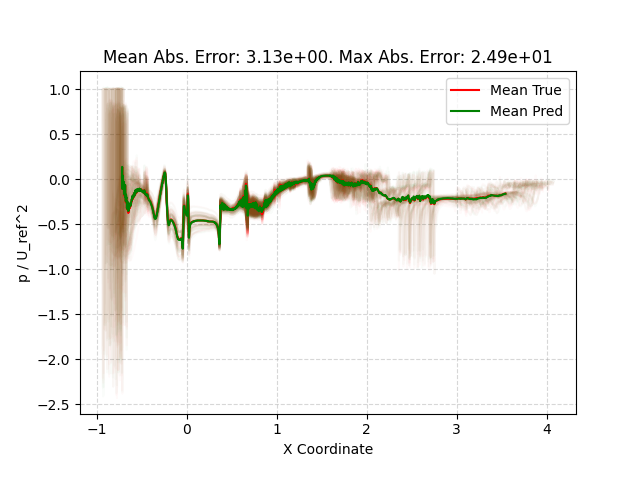}
         \label{fig:y equals x 7}
     \end{subfigure}
     \vfill
     \begin{subfigure}[b]{0.48\textwidth}
         \centering
         \subcaption{FIGConvNet Top Centerline}
         \includegraphics[trim={1cm 1cm 1cm 0.75cm},clip,width=\textwidth]{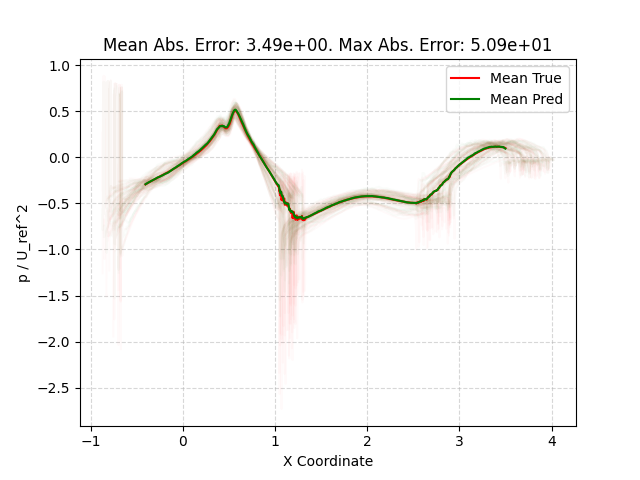}
         \label{fig:three sin x 7}
     \end{subfigure}
     \hfill
     \begin{subfigure}[b]{0.48\textwidth}
         \centering
         \subcaption{FIGConvNet Bottom Centerline}
         \includegraphics[trim={1cm 1cm 1cm 1cm},clip,width=\textwidth]{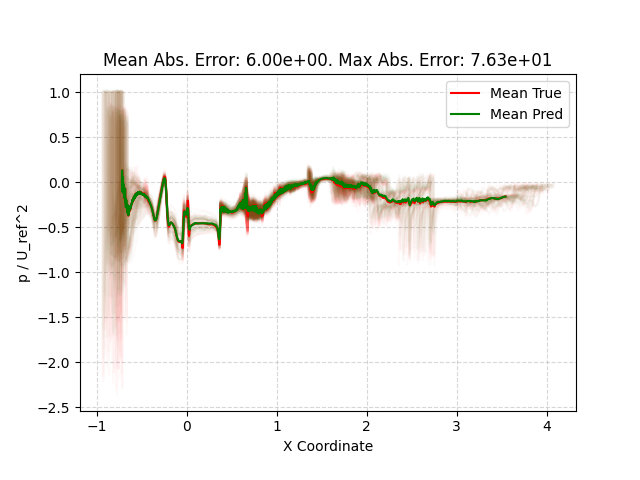}
         \label{fig:three sin x 8}
     \end{subfigure}
     \vfill
     \begin{subfigure}[b]{0.48\textwidth}
         \centering
         \subcaption{X-MeshGraphNet Top Centerline}
         \includegraphics[trim={1cm 1cm 1cm 0.75cm},clip,width=\textwidth]{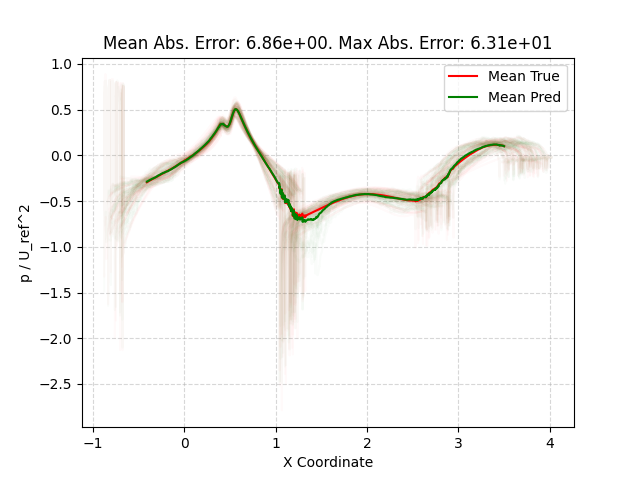}
         \label{fig:five over x 3}
     \end{subfigure}
     \hfill
     \begin{subfigure}[b]{0.48\textwidth}
         \centering
         \subcaption{X-MeshGraphNet Bottom Centerline}
         \includegraphics[trim={1cm 1cm 1cm 1cm},clip,width=\textwidth]{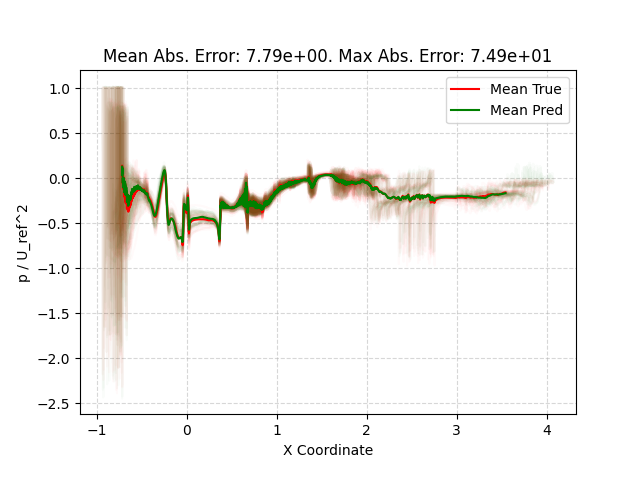}
         \label{fig:five over x 4}
     \end{subfigure}     
     
    \caption{Pressure predictions along vehicle centerline}
    \label{centerline-plots}
\end{figure}

\paragraph{Centerline plots} Figure~\ref{centerline-plots} shows the normalized pressure distribution along the vehicle centerline. The plot is split between the top centerline and bottom centerline for better visualization. The centerlines across all the samples are shown, with their mean displayed in darker color. All the models perform very well in this metric of comparison, with DoMINO performing slightly better than the other two, especially in capturing the pressure changes at sharp corners.

\paragraph{Surface contours}

\begin{figure}[ht!]
    \centering
    \begin{subfigure}[b]{0.47\textwidth}
        \centering
        \includegraphics[trim={1cm 2cm 1cm 5cm},clip,width=\textwidth]{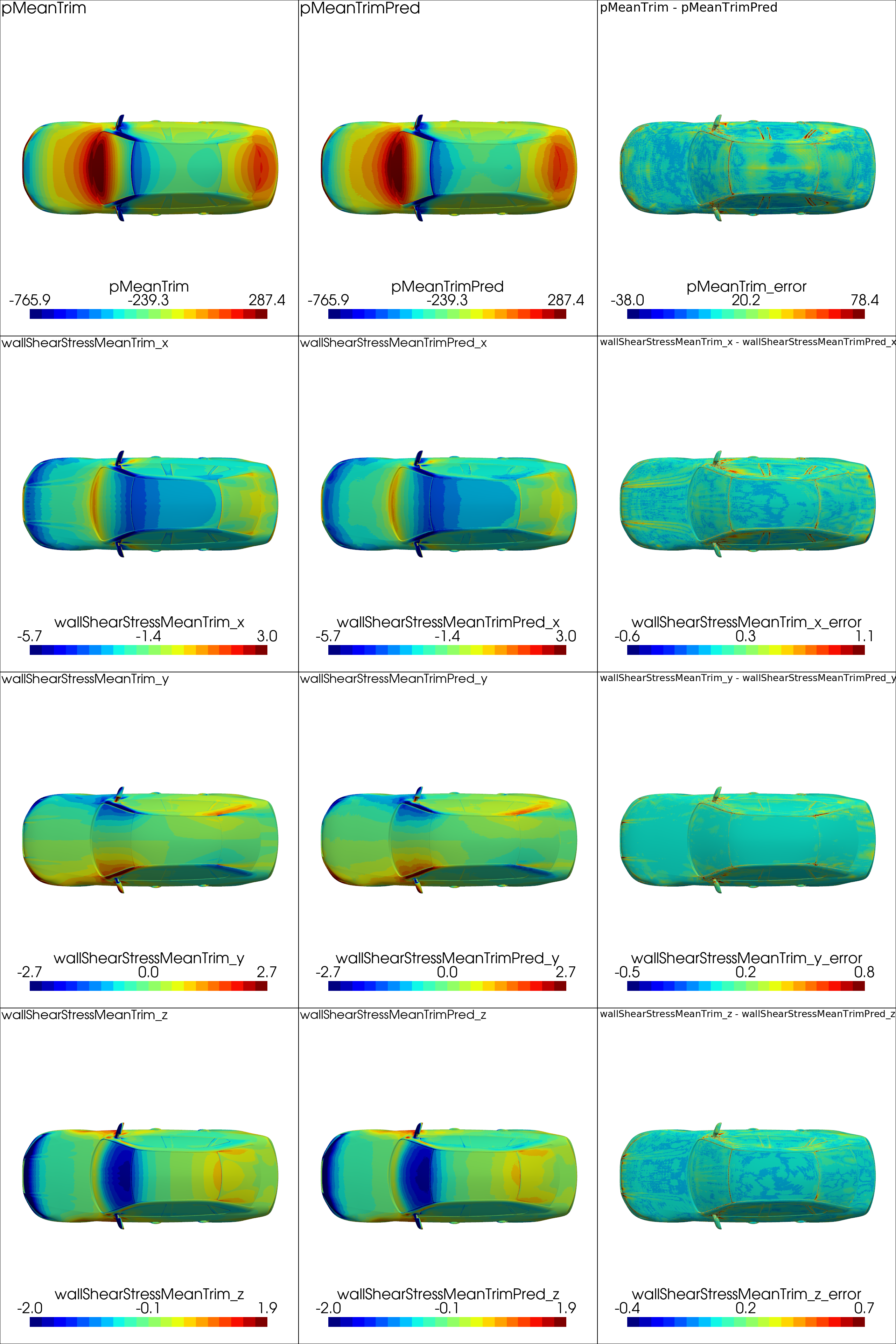}
        \caption{Test sample 419}
        \label{fig:y_equals_x}
    \end{subfigure}
    \hfill
    \begin{subfigure}[b]{0.47\textwidth}
        \centering
        \includegraphics[trim={1cm 2cm 1cm 5cm},clip,width=\textwidth]{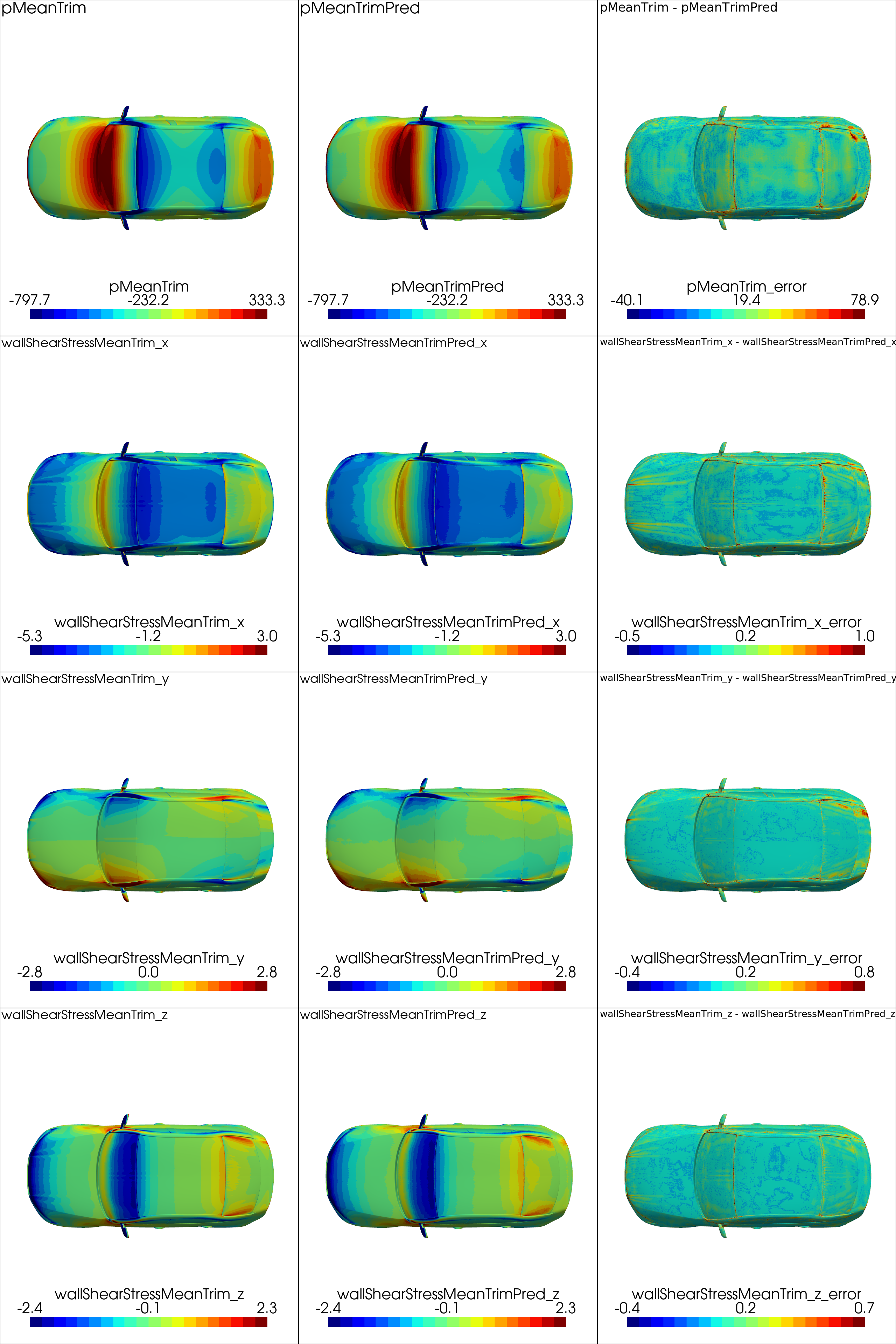}
        \caption{Test sample 439}
        \label{fig:three_sin_x}
    \end{subfigure}
    \caption{Surface contours along XY for DoMINO}
    \label{domino_surface_contour}
\end{figure}

\begin{figure}[ht!]
    \centering
    \begin{subfigure}[b]{0.47\textwidth}
        \centering
        \includegraphics[trim={1cm 2cm 1cm 5cm},clip,width=\textwidth]{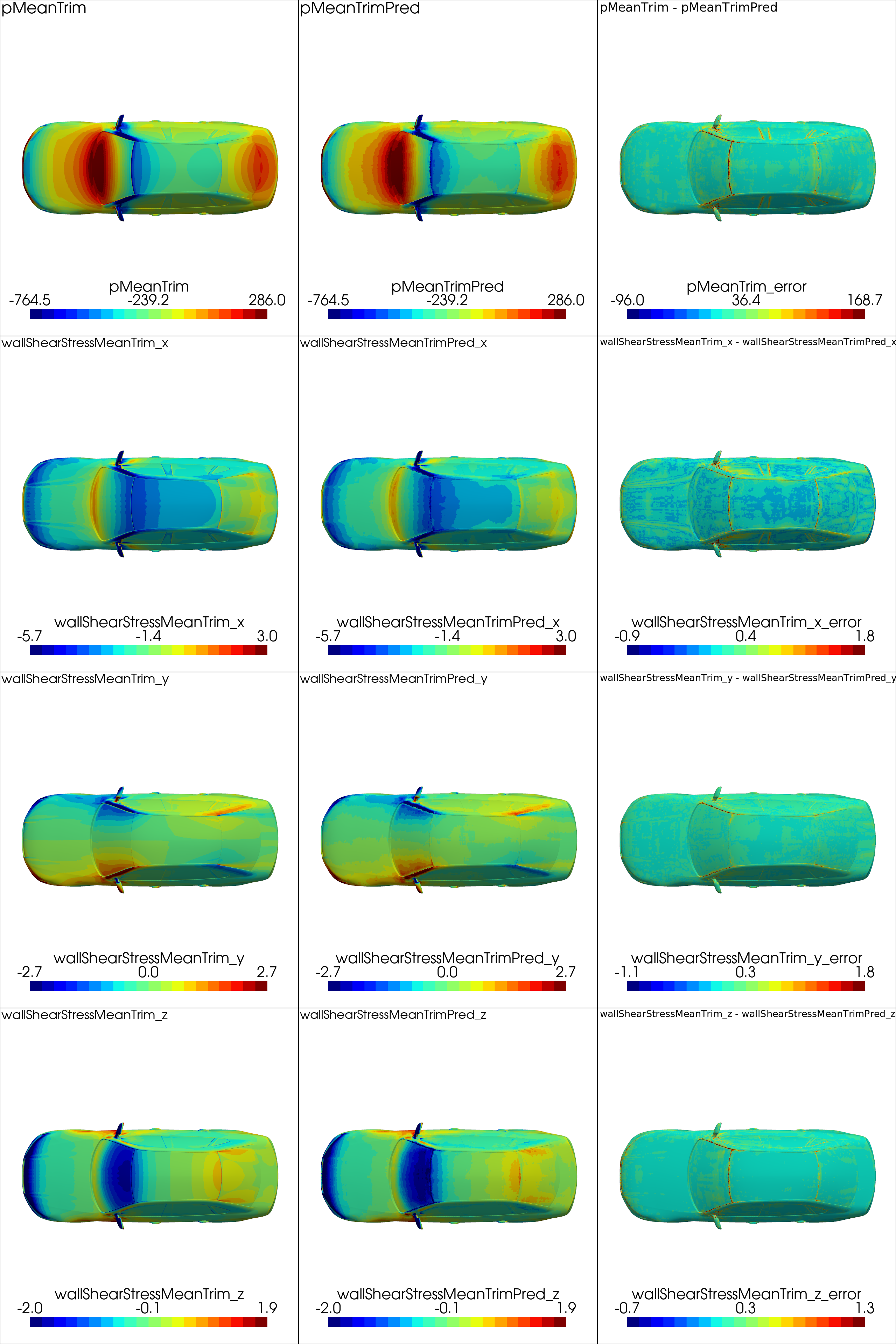}
        \caption{Test sample 419}
        \label{fig:y_equals_x_figcon}
    \end{subfigure}
    \hfill
    \begin{subfigure}[b]{0.47\textwidth}
        \centering
        \includegraphics[trim={1cm 2cm 1cm 5cm},clip,width=\textwidth]{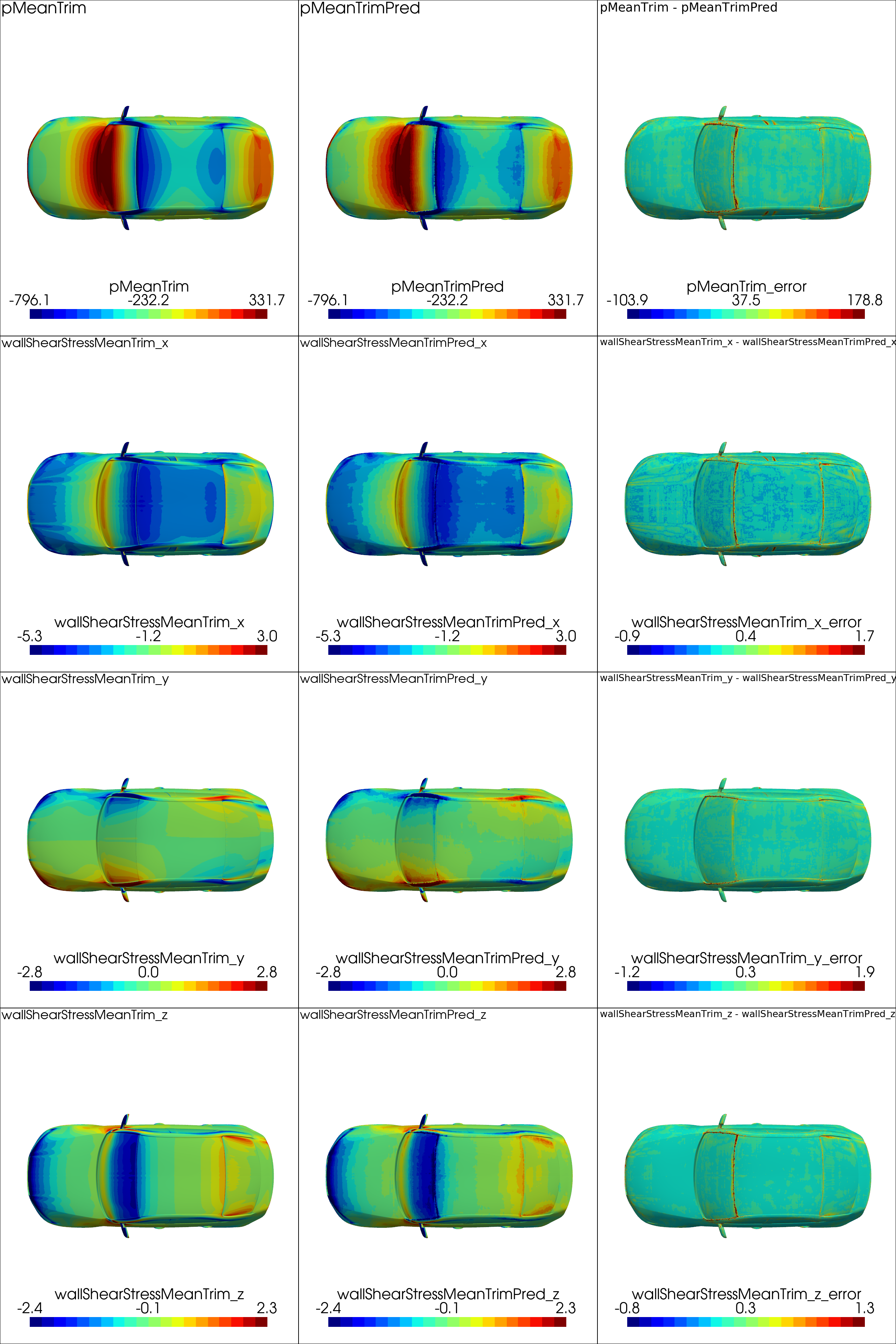}
        \caption{Test sample 439}
        \label{fig:three_sin_x_figcon}
    \end{subfigure}
    \caption{Surface contours along XY for FIGConvNet}
    \label{fig_surface_contour}
\end{figure}

\begin{figure}[ht!]
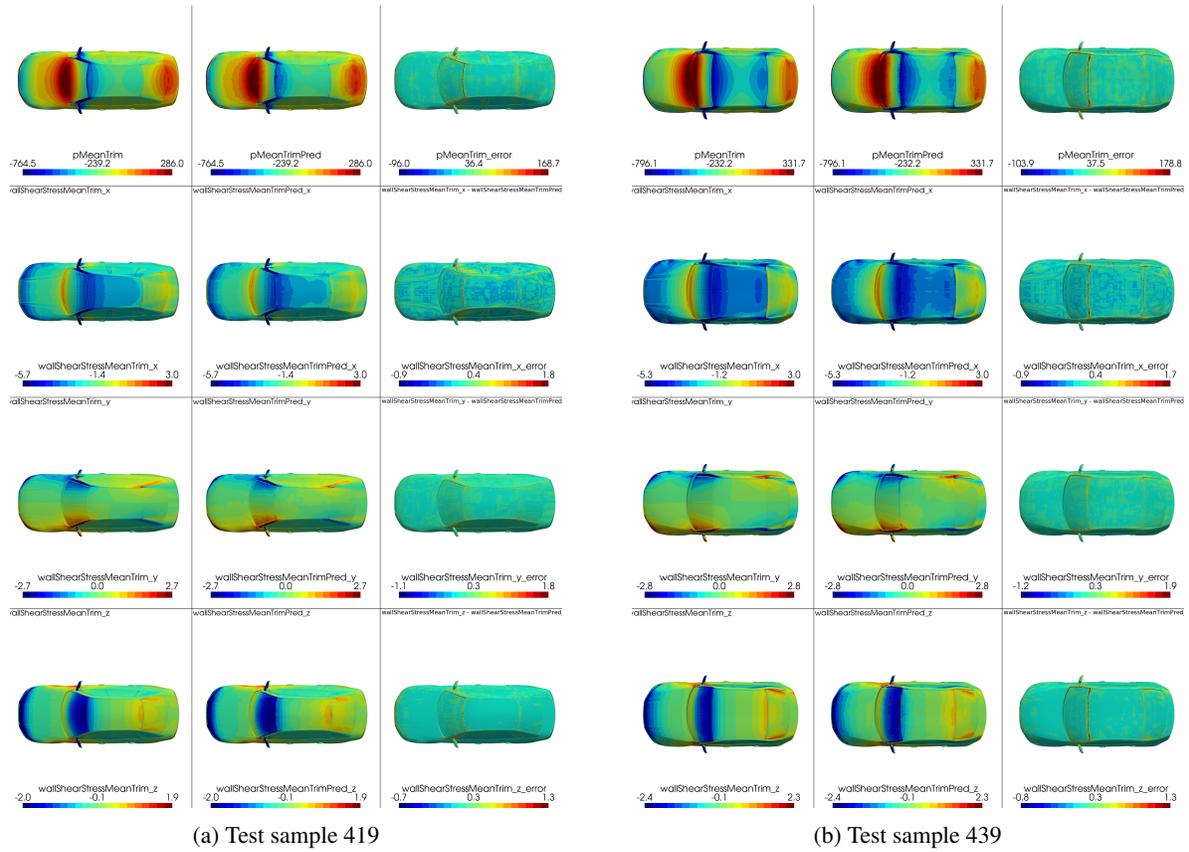

    \centering
    \begin{subfigure}[b]{0.47\textwidth}
        \centering
        \includegraphics[trim={1cm 2cm 1cm 5cm},clip,width=\textwidth]{figures/results/sample_outputs_figcon/compare_surface_xy_419.png}
        \caption{Test sample 419}
        \label{fig:sample419_xmesh}
    \end{subfigure}
    \hfill
    \begin{subfigure}[b]{0.47\textwidth}
        \centering
        \includegraphics[trim={1cm 2cm 1cm 5cm},clip,width=\textwidth]{figures/results/sample_outputs_figcon/compare_surface_xy_439.png}
        \caption{Test sample 439}
        \label{fig:sample439_xmesh}
    \end{subfigure}
    \caption{Surface contours along XY for X-MeshGraphNet}
    \label{xaero_surface_contour}
\end{figure}

We provide the surface contour visualizations in Figures~\ref{domino_surface_contour}, ~\ref{fig_surface_contour} and \ref{xaero_surface_contour} for completeness, however for inter-model comparison purposes, it is difficult to use this metric, and we recommend using the L2 errors / Area-weighted L2 errors which provide an quantitative way to compare the surface predictions.

\paragraph{Evaluation on uniform point clouds}

As mentioned in Section~\ref{sec:val-on-mesh-vs-pc}, we also evaluate the models based on the predictions made on a uniform point cloud of 10 million points on the surface. Table ~\ref{table:r2-scores-pc} \ref{table:trend-spearman-pc}, and ~\ref{table:trend-errors-pc} provide the comparisons of integrated drag and lift forces. When comparing the integrated forces, the model results are compared to the ground truth forces that are evaluated using the simulation mesh. FIGConvNet and DoMINO perform better than X-MeshGraphNet. The better performance of FIGConvNet and DoMINO is likely because of a loss term for the integrated quantities.



\begin{table}[H]
\centering
\caption{R2 scores evaluated using point cloud of 10 M points on validation samples (Higher is better)}
\label{table:r2-scores-pc}
\begin{tabular}{lrrr}
\toprule
 & X-MeshGraphNet & FIGConvNet & DoMINO \\
\midrule
R2 Drag & \cellcolor{blue0}0.85 & \cellcolor{blue19}0.97 & \cellcolor{blue18}0.97 \\
R2 Lift & \cellcolor{blue0}0.41 & \cellcolor{blue18}0.95 & \cellcolor{blue19}0.95 \\
\bottomrule
\end{tabular}
\end{table}

\begin{table}[H]
\centering
\caption{Trend analysis evaluated using point cloud of 10 M points (Spearman coefficient) on validation samples (Higher is better)}
\label{table:trend-spearman-pc}
\begin{tabular}{lrrr}
\toprule
 & X-MeshGraphNet & FIGConvNet & DoMINO \\
\midrule
Spearman Coeff. Drag & \cellcolor{blue0}0.92 & \cellcolor{blue18}0.98 & \cellcolor{blue19}0.99 \\
Spearman Coeff. Lift & \cellcolor{blue0}0.74 & \cellcolor{blue19}0.98 & \cellcolor{blue18}0.97 \\
\bottomrule
\end{tabular}
\end{table}

\begin{table}[H]
\centering
\caption{Trend analysis evaluated using point cloud of 10 M points (Errors) on validation samples (Lower is better)}
\label{table:trend-errors-pc}
\begin{tabular}{lrrr}
\toprule
 & X-MeshGraphNet & FIGConvNet & DoMINO \\
\midrule
Mean Abs. Error Drag & \cellcolor{blue0}21.61 & \cellcolor{blue19}9.02 & \cellcolor{blue18}9.33 \\
Max Abs. Error Drag & \cellcolor{blue0}74.83 & \cellcolor{blue19}29.77 & \cellcolor{blue18}30.56 \\
Mean Abs. Error Lift & \cellcolor{blue0}73.97 & \cellcolor{blue19}19.66 & \cellcolor{blue18}19.91 \\
Max Abs. Error Lift & \cellcolor{blue0}205.81 & \cellcolor{blue19}66.81 & \cellcolor{blue17}81.08 \\
\bottomrule
\end{tabular}
\end{table}

\subsubsection{Volume results}

\begin{figure}[ht!]
    \centering
    \begin{subfigure}[b]{0.47\textwidth}
        \centering
        \includegraphics[trim={1cm 2cm 1cm 5cm},clip,width=\textwidth]{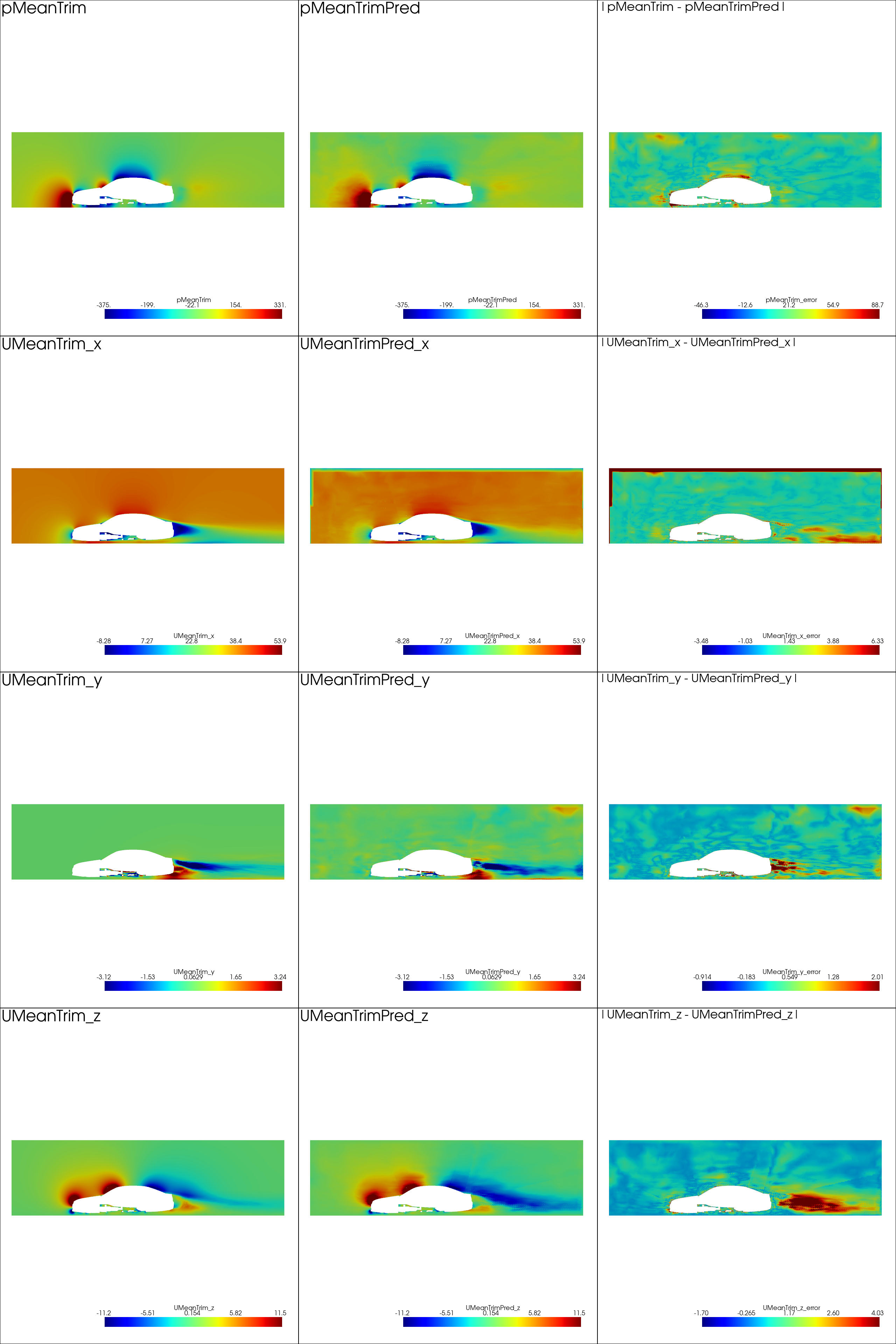}
        \caption{Test sample 419}
        \label{fig:domino_volume_419}
    \end{subfigure}
    \hfill
    \begin{subfigure}[b]{0.47\textwidth}
        \centering
        \includegraphics[trim={1cm 2cm 1cm 5cm},clip,width=\textwidth]{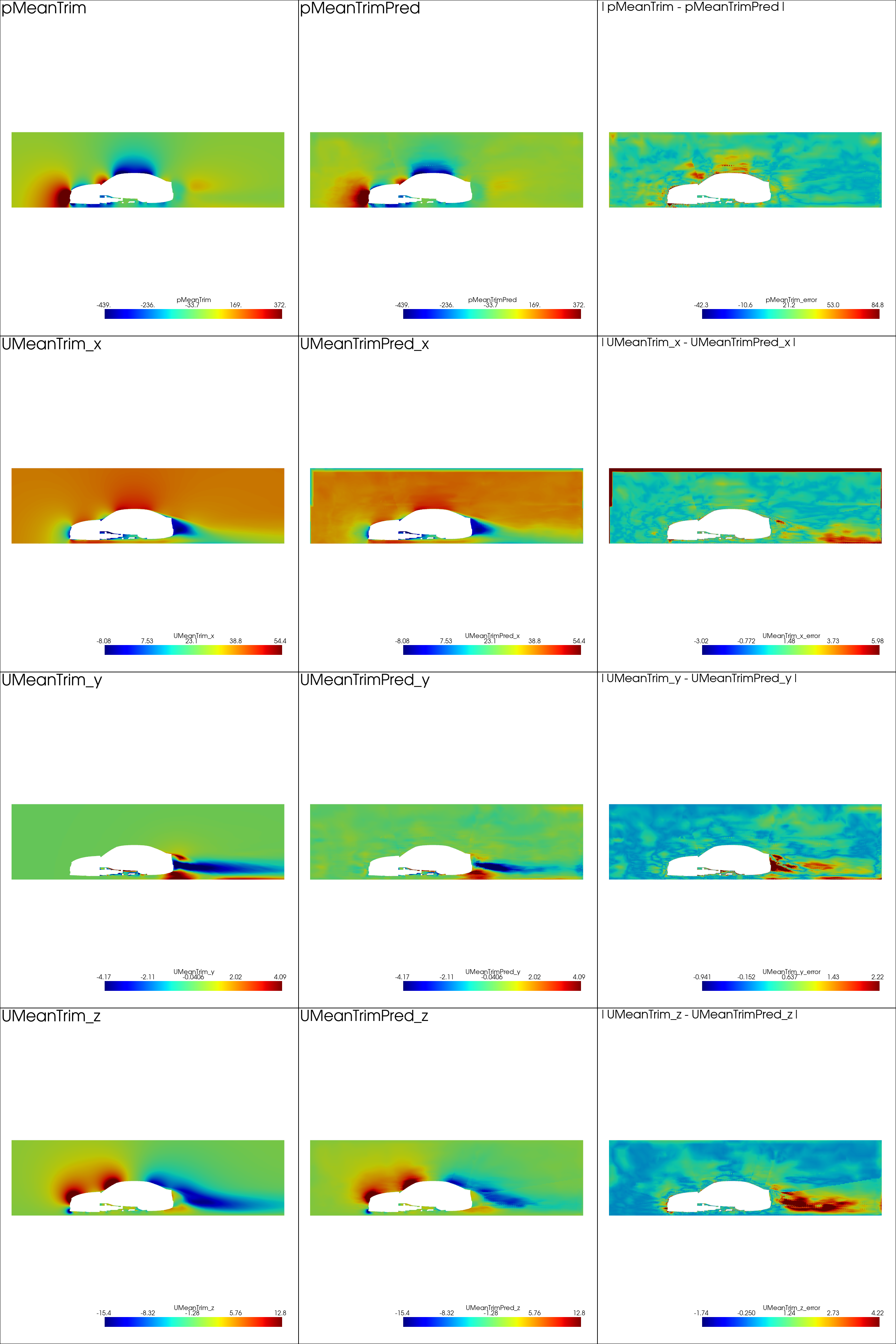}
        \caption{Test sample 439}
        \label{fig:domino_volume_439}
    \end{subfigure}
    \caption{Volume contours along XZ plane for DoMINO}
    \label{domino_volume_contour}
\end{figure}

\begin{figure}[ht!]
     \centering
     \begin{subfigure}[b]{0.4\textwidth}
         \centering
         \caption{Test sample 419}
         \includegraphics[trim={0.5cm 0.5cm 0.5cm 0.5cm},clip,width=\textwidth]{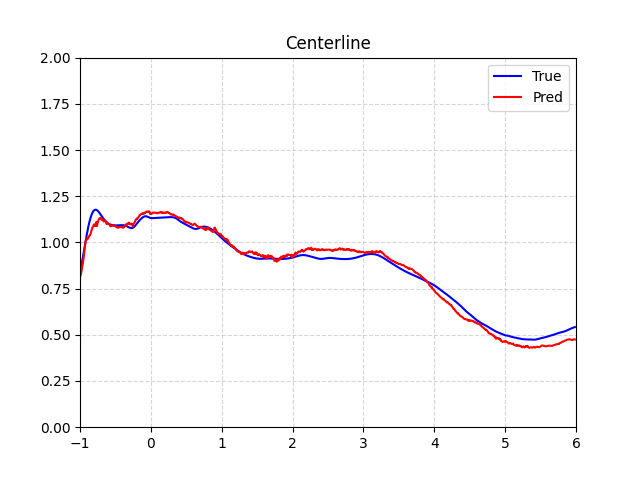}
         \label{fig:y equals x 8}
     \end{subfigure}
     \hfill
     \begin{subfigure}[b]{0.4\textwidth}
         \centering
         \caption{Test sample 439}
         \includegraphics[trim={0.5cm 0.5cm 0.5cm 0.5cm},clip,width=\textwidth]{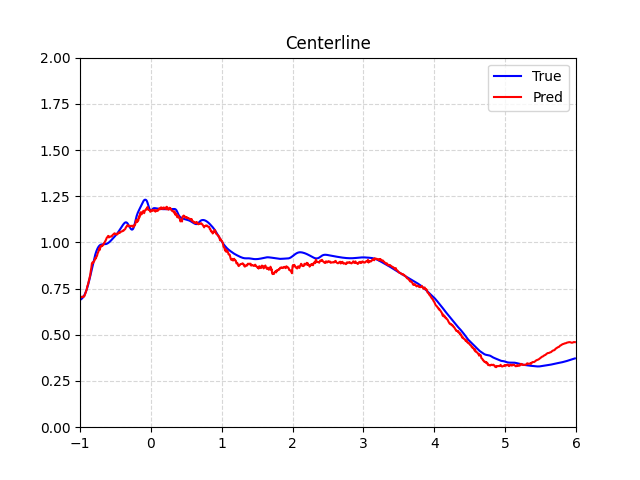}
         \label{fig:three sin x 1}
     \end{subfigure}
    \vfill
    \begin{subfigure}[b]{0.4\textwidth}
         \centering
         \includegraphics[trim={0.5cm 0.5cm 0.5cm 0.5cm},clip,width=\textwidth]{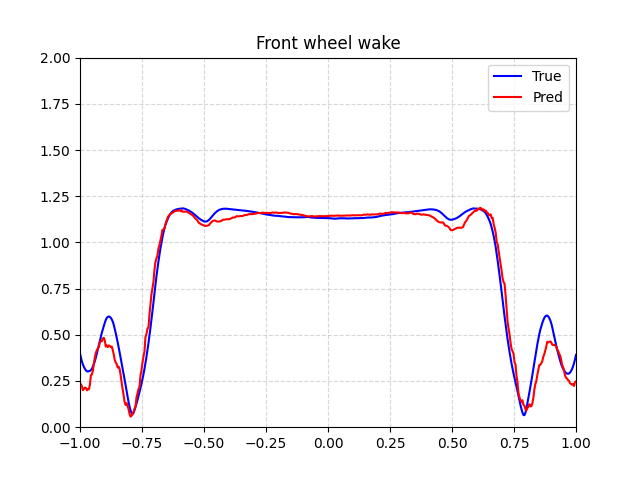}
         \label{fig:y equals x 9}
     \end{subfigure}
     \hfill
     \begin{subfigure}[b]{0.4\textwidth}
         \centering
         \includegraphics[trim={0.5cm 0.5cm 0.5cm 0.5cm},clip,width=\textwidth]{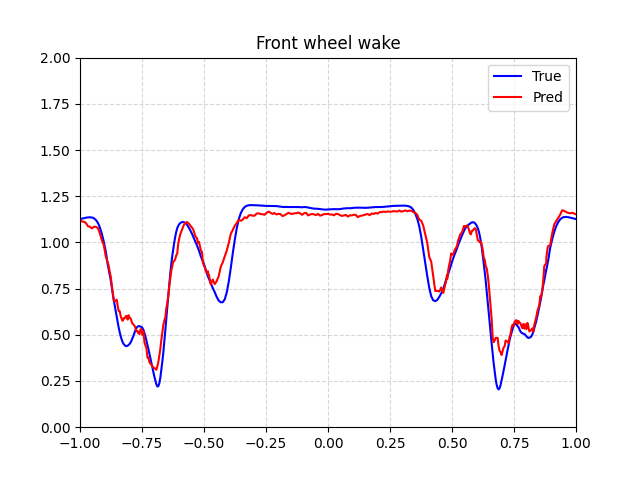}
         \label{fig:three sin x 2}
     \end{subfigure}
     \vfill
     \begin{subfigure}[b]{0.4\textwidth}
         \centering
         \includegraphics[trim={0.5cm 0.5cm 0.5cm 0.5cm},clip,width=\textwidth]{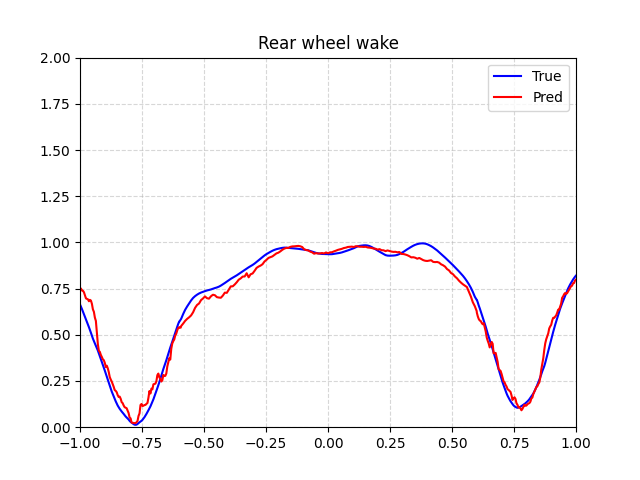}
         \label{fig:y equals x 1}
     \end{subfigure}
     \hfill
     \begin{subfigure}[b]{0.4\textwidth}
         \centering
         \includegraphics[trim={0.5cm 0.5cm 0.5cm 0.5cm},clip,width=\textwidth]{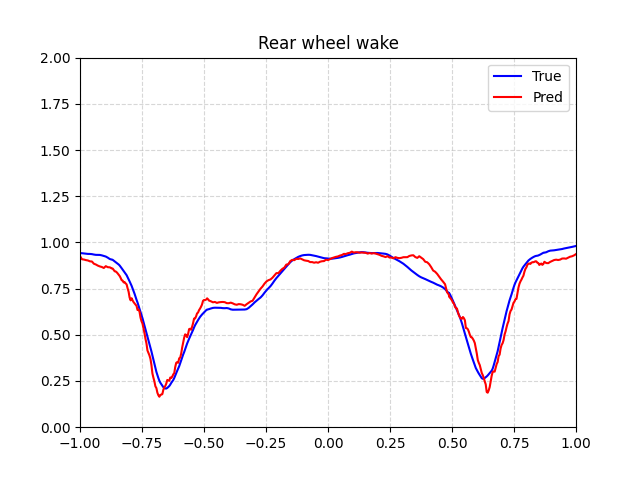}
         \label{fig:three sin x}
     \end{subfigure}
     
    \caption{Velocity along centerline (top), front wheel wake (middle) and rear wheel wake (bottom)}
    \label{wake-comparisons-1}
\end{figure}

\begin{figure}[ht!]
     \centering
     \begin{subfigure}[b]{0.4\textwidth}
         \centering
         \caption{Test sample 419}
         \includegraphics[trim={0.5cm 0.5cm 0.5cm 0.5cm},clip,width=\textwidth]{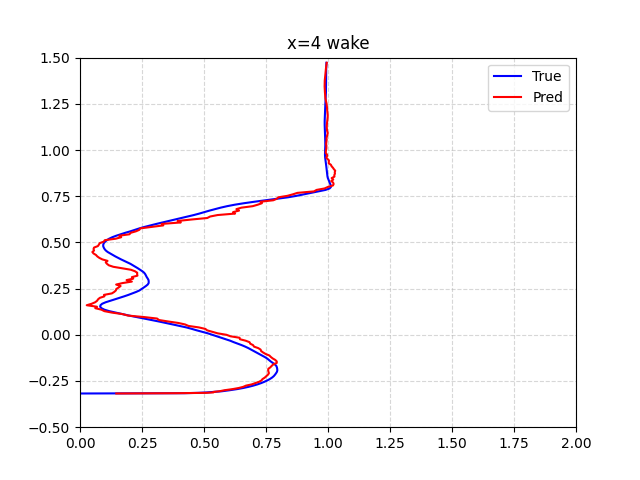}
         \label{fig:y equals x 2}
     \end{subfigure}
     \hfill
     \begin{subfigure}[b]{0.4\textwidth}
         \centering
         \includegraphics[trim={0.5cm 0.5cm 0.5cm 0.5cm},clip,width=\textwidth]{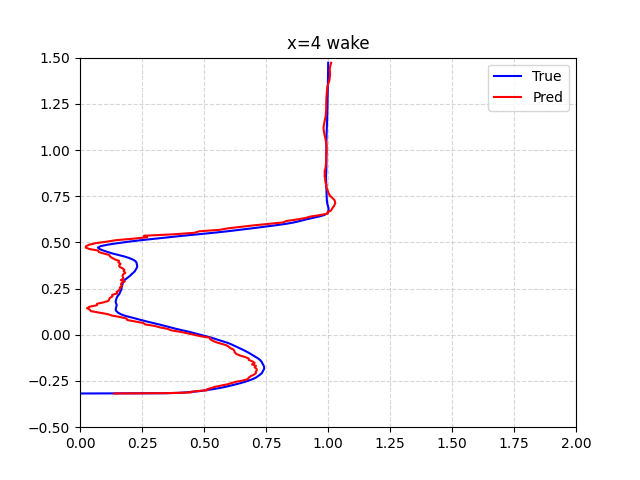}
         \label{fig:three sin x 3}
     \end{subfigure}
     \vfill
     \begin{subfigure}[b]{0.4\textwidth}
         \centering
         \caption{Test sample 439}
         \includegraphics[trim={0.5cm 0.5cm 0.5cm 0.5cm},clip,width=\textwidth]{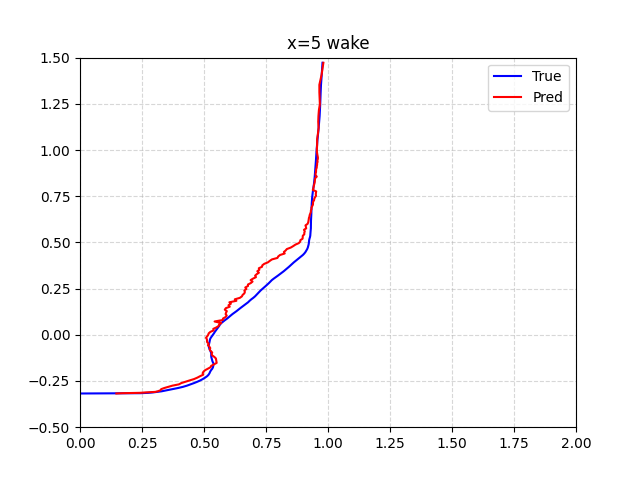}
         \label{fig:y equals x 3}
     \end{subfigure}
     \hfill
     \begin{subfigure}[b]{0.4\textwidth}
         \centering
         \includegraphics[trim={0.5cm 0.5cm 0.5cm 0.5cm},clip,width=\textwidth]{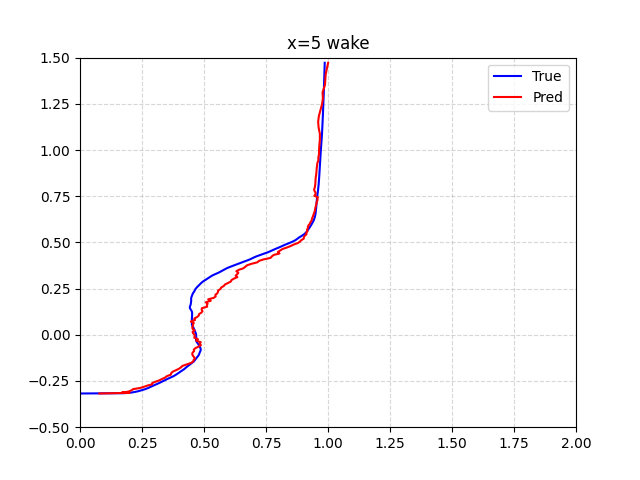}
         \label{fig:three sin x 4}
     \end{subfigure}
     
    \caption{Velocity along Z at x=4, y=0 (top) and x=5, y=0 (bottom)}
    \label{wake-comparisons-2}
\end{figure}

For volume comparisons, we only provide the results of DoMINO in Table~\ref{volume-table} and Figures~\ref{domino_volume_contour}, ~\ref{wake-comparisons-1}, and ~\ref{wake-comparisons-2}. Development of FIGConvNet for volume predictions is currently ongoing. X-MeshGraphNet provides volume prediction capabilities, but due to resource and time constraints, we have not fine-tuned a volume model for X-MeshGraphNet. Hence, for volume comparisons, we only include the results from the DoMINO model. However, similar to the surface benchmarking, the workflows from PhysicsNeMo-CFD can be used for making inter-model comparisons.

\begin{table}[H]
\centering
  \caption{Average volume error metrics on validation samples}
  \label{volume-table}
  \begin{tabular}{lr}
  \toprule
     & DoMINO \\
    \midrule
    X-Velocity & 0.0948 \\ 
    Y-Velocity & 0.1848 \\ 
    Z-Velocity & 0.2040 \\ 
    Pressure & 0.1042  \\ 
    \bottomrule
  \end{tabular}
\end{table}

\section{Conclusion and Future work} \label{conclusion}
In this work, we introduce a benchmarking framework in PhysicsNeMo-CFD that provides a consistent set of CFD specific metrics to enable development and thorough analysis of AI models. The benchmarking framework is highly customizable and applicable to a variety of external aerodynamics use cases, not limited to the datasets discussed in this study. We demonstrate the utility of this framework with three AI models, namely FIGConvNet, X-MeshGraphNet, and DoMINO. The benchmarking framework helped in validating the choices made in architecture, data representation, and loss functions—insights and significantly aided the development of these models. It was crucial throughout the development of these models, providing guidance in making key decisions that shaped their development. 

The landscape of Physics-AI is becoming increasingly rich with the availability of standardized datasets and benchmarks, similar to the advancements seen in domains like NLP and vision. Datasets such as ERA5 for weather and DrivAerML for external aerodynamics have become available in the open-source domain. While benchmarks like WeatherBench \cite{rasp2023weatherbench} and \href{https://github.com/NVIDIA/earth2studio}{Earth2Studio} have standardized the weather and climate domains, the field of CFD still lacks such standardized capabilities. PhysicsNeMo-CFD aims to bridge this gap and enable the Physics-AI community to push the boundaries of developing accurate and generalizable model architecture.

Future work will focus on adding more domain-specific metrics for comparison and improving its usability to enable domain scientists to conduct more scientific exploration with trained models. This includes features like solver initialization (already a part of PhysicsNeMo-CFD), design sensitivity analysis, and enhancements in physics-based testing and uncertainty quantification. Additionally, we aim to extend PhysicsNeMo-CFD to provide benchmarks for other standardized Physics-ML CFD use cases, such as turbulence modeling. 

\section{Acknowledgments}
The authors would like to thank Dr. Faez Ahmed and Mohamed Elrefaie from Massachusetts Institute of Technology for their valuable guidance and insightful discussions throughout this work.

%

%

%

%

\FloatBarrier

\bibliographystyle{ieeetr} 
\bibliography{main}

\end{document}